\documentclass[letterpaper]{article} 

\ifdefined\aaaianonymous
    \usepackage[submission]{aaai2026}  
\else
    \usepackage{aaai2026}              
\fi

\usepackage{makecell}
\usepackage{amsmath}
\usepackage{amssymb}
\usepackage{booktabs}   
\usepackage{multirow}   
\usepackage{graphicx}   
\usepackage[table]{xcolor}
\usepackage{arydshln}

\usepackage{times}  
\usepackage{helvet}  
\usepackage{courier}  
\usepackage[hyphens]{url}  
\usepackage{graphicx} 
\usepackage{natbib}  
\usepackage{caption} 
\usepackage{algorithm}
\usepackage{algorithmic}

\usepackage{newfloat}
\usepackage{listings}
\DeclareCaptionStyle{ruled}{labelfont=normalfont,labelsep=colon,strut=off} 
\floatstyle{ruled}
\newfloat{listing}{tb}{lst}{}
\floatname{listing}{Listing}

\ifdefined\aaaianonymous
    \title{AAAI 2026 Supplementary Material\\Anonymous Submission}
\else
    \title{Decomposing and Composing: Towards Efficient Vision-Language Continual Learning via Rank-1 Expert Pool in a Single LoRA}
\fi

\ifdefined\aaaianonymous
\author{
    Anonymous Submission
}
\affiliations{
}
\else
\author{
Zhan Fa\textsuperscript{1}, Yue Duan\textsuperscript{1}, Jian Zhang\textsuperscript{1}, Lei Qi\textsuperscript{2}, Wanqi Yang\textsuperscript{3}, Yinghuan Shi\textsuperscript{1}\thanks{Corresponding author}\\
}
\affiliations{
\textsuperscript{1}National Key Laboratory for Novel Software Technology, Nanjing University, China\\
\textsuperscript{2}School of Computer Science and Engineering, Southeast University, China\\
\textsuperscript{3}School of Computer and Electronic Information, Nanjing Normal University, China\\
\{fazhan, yueduan, zhangjian7369\}@smail.nju.edu.cn, qilei@seu.edu.cn, yangwq@njnu.edu.cn, syh@nju.edu.cn
}

\fi
\begin{document}

\maketitle

\begin{abstract}
Continual learning (CL) in vision-language models (VLMs) faces significant challenges in improving task adaptation and avoiding catastrophic forgetting. Existing methods usually have heavy inference burden or rely on external knowledge, while Low-Rank Adaptation (LoRA) has shown potential in reducing these issues by enabling parameter-efficient tuning. However, considering directly using LoRA to alleviate the catastrophic forgetting problem is non-trivial, we introduce a novel framework that restructures a single LoRA module as a \textbf{decomposable} Rank-1 Expert Pool. Our method learns to dynamically \textbf{compose} a sparse, task-specific update by selecting from this expert pool, guided by the semantics of the \texttt{[CLS]} token.
In addition, we propose an \textbf{Activation-Guided Orthogonal (AGO) loss} that orthogonalizes critical parts of LoRA weights across tasks. This sparse composition and orthogonalization enable fewer parameter updates, resulting in domain-aware learning while minimizing inter-task interference and maintaining downstream task performance. Extensive experiments across multiple settings demonstrate state-of-the-art results in all metrics, surpassing zero-shot upper bounds in generalization. Notably, it reduces trainable parameters by 96.7\% compared to the baseline method, eliminating reliance on external datasets or task-ID discriminators. The merged LoRAs retain less weights and incur no inference latency, making our method computationally lightweight. 
\end{abstract}
\begin{links}
    \link{Code}{https://github.com/Fazhan-cs/DAC}
\end{links}
\begin{figure}[h]

  \centering
  \includegraphics[width=0.95\linewidth]{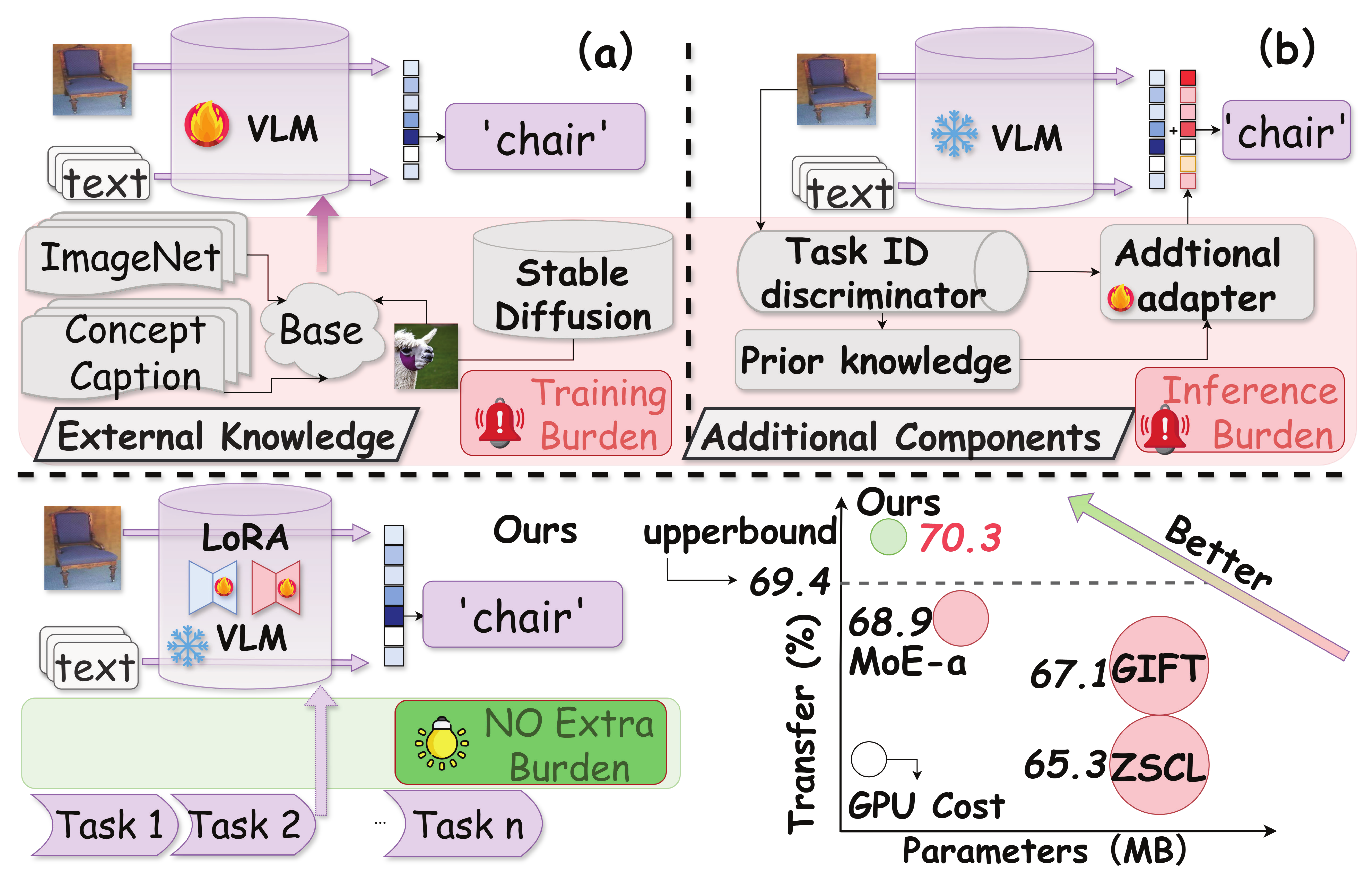}

  \caption{ Advantages of our method over previous works: (a) ZSCL and GIFT introduce large external datasets or synthetic images from generative models for model regularization. (b) MoE-a  and RAIL  introduce additional components and utilize prior knowledge during testing. Our method uses \textbf{NO} external data knowledge and introduces \textbf{NO} additional burden. Meanwhile, it has fewer training parameters and lower GPU cost, with Transfer, a metric to measure the generalization ability of VLMs, exceeding the upper bound of the original CLIP zero-shot performance.
  }

  \label{fig:intro}

\end{figure}
\section{Introduction}

Vision-Language Models (VLMs) are vital for real-world multi-domain tasks as they possess the ability to address the challenging problem of aligning visual and language modalities  \cite{parelli2023clip,antol2015vqa,hong2024navigating}. In real-world scenarios, continual learning (CL) in VLMs is imperative due to the dynamic and evolving nature of real-world data, and systems need to adapt continuously to new information  \cite{zheng2023preventing,jha2024clap4clip}. Leveraging pre-trained VLM like CLIP  \cite{radford2021learning}, recent CL advancements enhance downstream task performance via continual fine-tuning  \cite{jha2024clap4clip,thengane2022clip,lee2023pre}. In this paper, we center on multi-domain continual learning of VLMs, aiming to boost downstream tasks and maintain the transfer ability of pre-trained CLIP models. Many works have explored the learning and forgetting problems of VLMs  \cite{wu2025synthetic,yu2024boosting,xu2024advancing}. Previous works, as depicted in Figure \ref{fig:intro}, have tangible limitations in the training and inference stages of continual learning:
(a) \textbf{Training burden}: Works such as ZSCL  \cite{zheng2023preventing} and GIFT  \cite{wu2025synthetic} heavily rely on external data sources. ZSCL uses over 100K images in large-scale datasets like ImageNet  \cite{deng2009imagenet} for model regularization, while GIFT utilizes Stable Diffusion  \cite{rombach2021highresolution} to generate synthetic images for over 10 hours for data replay.
(b) \textbf{Inference burden}: Methods like MoE-a   \cite{yu2024boosting} and RAIL  \cite{xu2024advancing} introduce additional complexity during inference. They add adapters to the pre-trained model, and the size of these additional components increases linearly with task numbers, i.e., 11 tasks here, bringing a great inference overhead. Moreover, they obtain prior knowledge such as task-id through discriminators during inference, which is difficult to obtain under real-world scenarios. 

Moreover, some previous works have attempted to use LoRA to address continual learning problems  \cite{wei2024online,wang2023orthogonal,yang2024parameter}, often by isolating training stage updates into multiple separate low-rank matrices. However, directly relying on LoRA to alleviate catastrophic forgetting is non-trivial. As previous studies have shown  \cite{jiang2025fine,gekhman2024does}, LoRA still suffers from redundant parameter updates despite its low-rank characteristics. While many LoRA-based studies have focused on static rank pruning to reduce this redundancy  \cite{jiang2025fine,zhang2023increlora,valipour2022dylora,ding2023sparse,meng2024pissa}, such approaches are not suitable for dynamic data injection and changing task domains in continual learning. To address this, we introduce a dynamic learning approach. We innovatively restructure a single LoRA module as a \textbf{decomposable} Rank-1 Expert Pool. The key insight behind our method is the equivalence that \textbf{a single LoRA with rank $r$ is equivalent to using $r$ LoRAs with rank $1$}. This decomposition allows us to solve the parameter redundancy problem from a new perspective: instead of updating the entire low-rank matrix, we can dynamically \textbf{compose} a sparse, task-specific update by selecting only the most relevant experts from the pool. This process makes the parameter updates more targeted and efficient, significantly reducing unnecessary interference and leading to better performance. Our framework exploits this efficient update mechanism while retaining the core advantages of LoRA. The training process remains highly parameter-efficient, and after each task, the composed sparse LoRA weights are merged back into the original model. This ensures our method introduces no extra components or latency during inference, creating a truly lightweight and effective solution for continual learning. To implement the dynamic selection, we then employ the semantic-rich \texttt{[CLS]} token  \cite{liang2022not,wang2024cls} to guide a lightweight router, enabling the selection of appropriate experts to adapt to different task domains.

In addition, considering that the orthogonality of parameter update directions is helpful for isolating task optimization objectives  \cite{wang2023orthogonal,yang2024parameter,feng2025omoe}, we innovatively propose an \textbf{Activation-Guided Orthogonal (AGO) loss}. Our approach uses the expert activation frequency recorded during our dynamic composition process. When updating for each task, we leverage this pre-recorded information to calculate the orthogonal loss with all previous trained LoRAs. Unlike previous works that use additional components to determine domain information, our orthogonal strategy isolates domains in the parameter space, considering that LoRA parameters are not merely numerical adjustments but encapsulate crucial model update directions  \cite{wang2023orthogonal}. Therefore, by directly zeroing the weights of low-activation-frequency experts, the task can focus on being orthogonal to the crucial parts of previous tasks, without causing excessive interference to the current task. This can avoid the problem of parameter collision  \cite{yang2024parameter} to a certain extent.

In general, the contributions of our work are as follows:
\begin{itemize}
    \item We systematically analyze the training and inference burdens in prior VLM continual learning works and design an efficient framework to address them.
    \item We propose a novel method that restructures a single LoRA module as a \textbf{decomposable} Rank-1 Expert Pool, allowing for the dynamic \textbf{composition} of sparse, task-specific subspaces and reducing critical burden.
    \item We design a synergistic \textbf{Activation-Guided Orthogonal (AGO) loss} that leverages expert activation frequencies to reduce inter-task interference precisely.
\end{itemize}
In diverse experimental settings, despite reducing training parameters by \textbf{96.7\%} compared to the baseline (570.76 MB vs. 18.99 MB), our method achieves state-of-the-art (SOTA) results across all metrics, with the transfer metric even exceeding CLIP's zero-shot performance by 0.9\%.

\begin{figure*}
  \centering
  \includegraphics[width=1\linewidth]{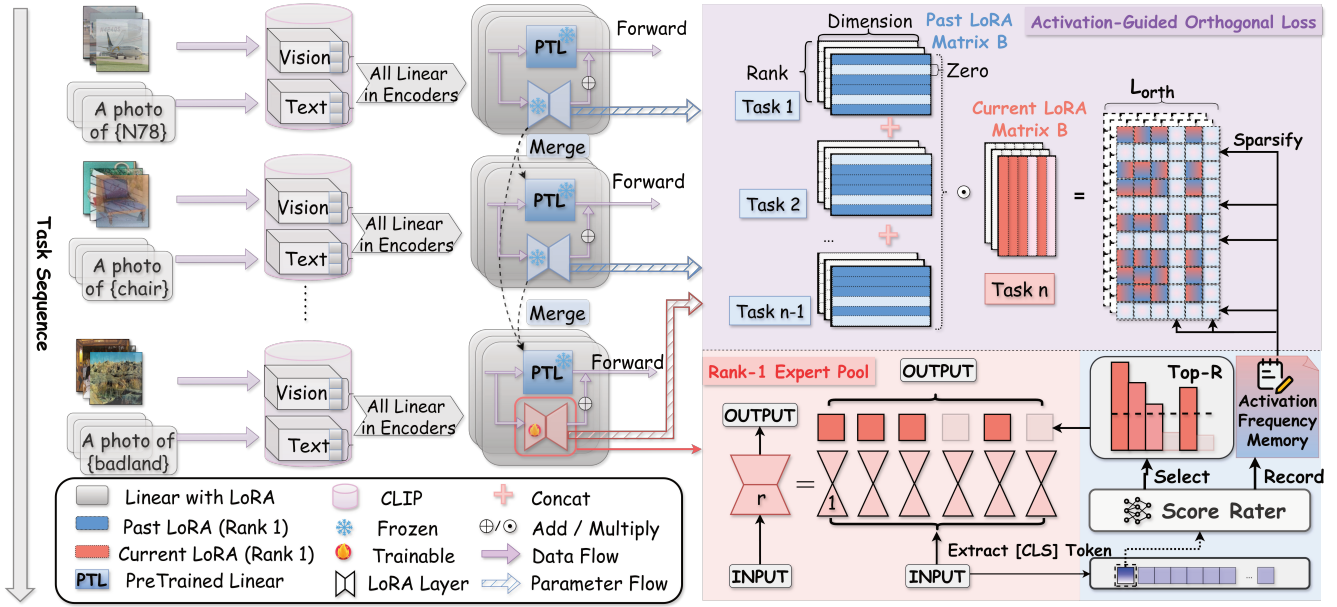}

  \caption{ The overall framework of our proposed method. LoRA is configured for all linear layers of the CLIP text and image encoder transformers, with the original parameters frozen. Each LoRA module is treated as a \textbf{Rank-1 Expert Pool}. For each input, the \texttt{[CLS]} token is extracted to guide a router that composes a sparse update by selecting critical experts from this pool. During training, an \textbf{Activation-Guided Orthogonal loss} is calculated between the current LoRA and those from past tasks. After training, the composed LoRA weights are merged back into the original model for zero-overhead inference. }
  \label{fig:overview}
\end{figure*}
\section{Related Works}

Existing continual learning methods primarily focus on solving Class Incremental Learning (CIL) or Task Incremental Learning (TIL) based on the domain changes of data 
 \cite{rebuffi2017icarl,agarwal2022semantics,wortsman2022robust,ding2022don}. Traditional continual learning methods encompass replay-based methods  \cite{rebuffi2017icarl,lavda2018continual}, distillation-based methods  \cite{ding2022don,li2017learning}, regularization-based methods  \cite{wortsman2022robust,aljundi2018memory}, and architecture-based methods  \cite{li2019learn,douillard2022dytox}. Here we focus on the setting of Multi-Task Incremental Learning (MTIL)  \cite{zheng2023preventing,yu2024boosting}, which enables VLMs to sequentially learn in different domains while retaining the pre-trained generalization ability for previously seen tasks. Recent works have focused on the continual fine-tuning of pre-trained VLMs: ZSCL  \cite{zheng2023preventing} adds a regularization term to the cross-entropy loss function to penalize changes in model parameters or feature space and regularize the parameter space using a large-scale reference dataset. MoE-a  \cite{yu2024boosting} cooperates the pre-trained CLIP with a Mixture of Experts (MoE) adapters and uses the reference dataset to discriminate task-ID, enabling the model to distinguish between unseen and seen tasks. Moreover, RAIL  \cite{xu2024advancing} uses the principle of ridge regression to add an additional high-dimensional classification adapter to CLIP and consolidates the learned knowledge using task IDs. GIFT  \cite{wu2025synthetic} uses a pre-trained diffusion model to generate replay images for the text inputs of past tasks to enhance memory.

These methods either introduce additional model components and datasets or require a large amount of prior knowledge for inference, which brings great training and inference burden. However, with the rapid development of LoRA in the era of pre-trained models  \cite{meng2024pissa,mao2024dora,zhang2023adalora,yang2024corda,ren2024analyzing}, we leverage its principles in continual learning scenarios.  Considering that LoRA can fine-tune VLM in a low-rank characteristic and merge the weights back into the original weights when training is completed, our method introduces no additional burdens during training and testing, making it more lightweight and more suitable for practical task scenarios. Furthermore, our method differs from previous work that treats multiple parallel LoRAs as individual experts in a Mixture of Experts (MoE) framework  \cite{liu2024moe,gao2024higher,dou2023loramoe}. Rather than adding more LoRA modules and increasing parameter counts, our approach restructures a single LoRA as a \textbf{decomposable} Rank-1 Expert Pool. We then use the semantics of the \texttt{[CLS]} token to dynamically \textbf{compose} sparse, task-specific subspaces from this internal expert pool. This provides a more fine-grained and parameter-efficient mechanism for continual learning.

\textbf{Remarks.} To sum up, while existing continual learning methods for VLMs are limited by their training and inference burdens, our work offers a significantly more efficient approach, by restructuring a single LoRA module as a decomposable Rank-1 Expert Pool that are dynamically composed for each task. This avoids the parameter and computational overhead caused by combining multiple, separate LoRA modules, offering a truly lightweight solution.

\section{Method}

\subsection{Problem Definition}

Following the standard setup for Vision-Language Models (VLMs) like CLIP  \cite{radford2021learning}, our model consists of an image encoder $f_{\theta}$ and a text encoder $g_{\psi}$ implemented as transformers. Classification is performed by calculating the cosine similarity between the image embeddings $\mathbf{z}^V = f_{\theta}(\mathbf{x})$ and text embeddings $\mathbf{z}^T = g_{\psi}(\mathbf{t})$. During inference, the probability of classifying the image $\mathbf{x}$ into class $y_i \in \{1,\ldots,\mathcal{C}\}$ is calculated as $p(y_i|\mathbf{x}) = \frac{\exp(\text{sim}(\mathbf{z}^V, \mathbf{z}_i^T))}{\sum_{c = 1}^{C}\exp(\text{sim}(\mathbf{z}^V, \mathbf{z}_c^T))}$, where $\text{sim}(\cdot)$ represents a cosine similarity metric.

In this work, we first evaluate our method on the multi-domain task-incremental learning (MTIL) scenario, a classic benchmark for multi-task and multi-domain continual learning  \cite{zheng2023preventing, yu2024boosting}. In this setting, the model learns sequentially from a series of 11 tasks. For each task $t$, the dataset is represented as $\mathcal{D}^t = \{(\mathbf{x}_i^t, y_i^t)\}_{i = 1}^{N^t}$, where $\mathbf{x}_i^t \in \mathbb{R}^{H\times W\times C}$ is the input image, $y_i^t \in \mathcal{C}^t$ is its corresponding class label, and $N^t$ is the number of samples in task $t$, respectively. The class set $\mathcal{C}^t = \{y_j^t\}_{j = 1}^{M^t}$ consists of the class names in task $t$, with a total of $M^t$ classes. Additionally, to demonstrate the robustness and generalization of our approach, we adopt another cross-domain task-agnostic incremental learning (X-TAIL) setting  \cite{xu2024advancing}. This benchmark is particularly challenging because it introduces task-agnostic settings during training, making it suitable for evaluating model generalization under realistic conditions.

\subsection{Overview Framework}
The core design of our framework is to achieve efficient continual learning by intelligently managing parameter updates within a single module, thereby avoiding common dependencies on external data or inference overhead. To this end, we restructure a single LoRA module as a \textbf{decomposable} Rank-1 Expert Pool. Guided by the semantics of the \texttt{[CLS]} token, our method learns to dynamically \textbf{compose} a sparse, task-specific update by selecting from this expert pool. To complement this sparse composition, we introduce an \textbf{Activation-Guided Orthogonal (AGO) loss} that minimizes parameter collision  \cite{yang2024parameter} by isolating critical updates in the existing parameter space, ensuring efficient learning without complex architectural burdens.

The overall framework of our method is shown in Figure~\ref{fig:overview}. We freeze the original weights of CLIP and add LoRA layers to all linear layers in the multi-head attention and MLP blocks of the Image and Text encoders. After the training for each task is completed, the composed sparsified LoRA weights are merged back into the original CLIP weights to ensure no additional inference burden is introduced. This process is repeated for each subsequent task, and for each task, only the LoRA weights and frequency information need to be saved. Except for the first task, which has no ``Past LoRA'' currently, the AGO loss is applied in subsequent training to isolate the parameter optimization direction between the current and previous tasks' LoRAs, enhancing the model's adaptability to new tasks without being overly affected by previous parameter updates.

\subsection{Dynamic Composition from a Rank-1 Expert Pool}
\begin{figure}[t]
  \centering
  \includegraphics[width=1.0\linewidth]{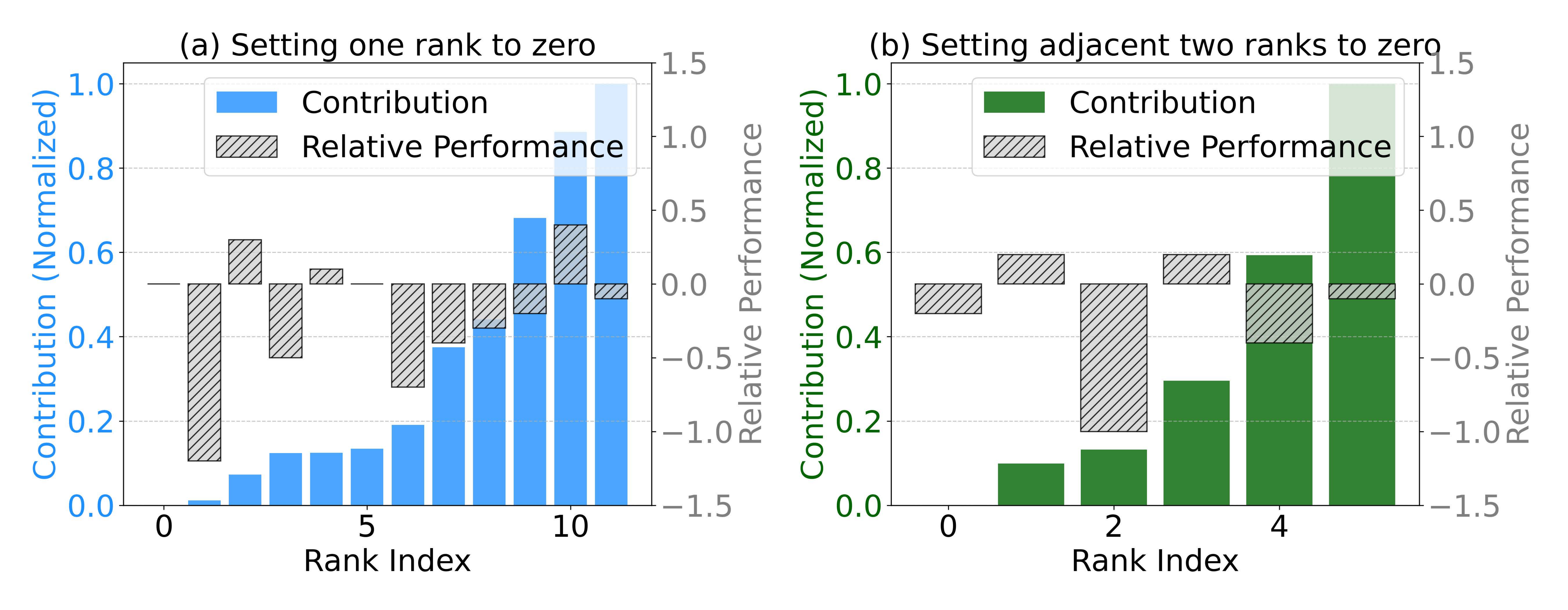}
  
  \caption{Comparison of the Average metric between setting no rank to zero and setting either one rank (a) or two adjacent ranks (b) to zero during the merging process. The rank is sorted in ascending order of Frobenius-norm: The contributions of each rank increase from left to right in sequence.
  The figure indicates that a higher contribution does not necessarily lead to performance improvement and shows ranks with low contribution may also play a crucial role in the task.
  }
  
  \label{fig:3.3}
\end{figure}

Many previous works fine-tuning with LoRA have introduced static metrics for sparsity and pruning  \cite{meng2024pissa,zhang2023adalora}. For example, DoRA  \cite{mao2024dora} uses the Frobenius norm to measure rank importance. However, we argue that such static analyses are ill-suited for the dynamic nature of continual learning. As shown in Figure \ref{fig:3.3}, the static importance of a rank does not reliably predict its contribution to performance in a dynamic setting. This finding motivates our dynamic approach: instead of treating LoRA as a monolithic block to be pruned, we view it as a pool of fine-grained, rank-1 experts that can be dynamically composed based on different task demands.

Our approach begins by decomposing a rank-$r$ LoRA matrix $\Delta W$. Following  \cite{hu2022lora}, $\Delta W = \mathbf{B}\mathbf{A}$, where $\mathbf{B} \in \mathbb{R}^{d\times r}$ and $\mathbf{A} \in \mathbb{R}^{r\times d}$. This can be expressed as the sum of $r$ rank-1 matrices:
\setlength{\arraycolsep}{0.4pt}  
\begin{equation}
\Delta W = \begin{bmatrix} \mathbf{b}_1&\mathbf{b}_2&\cdots&\mathbf{b}_r \end{bmatrix}\! \begin{bmatrix} \mathbf{a}_1^\top\ \\ \mathbf{a}_2^\top \\ \vdots \\ \mathbf{a}_r^\top \end{bmatrix}\!=\sum_{i = 1}^{r}\mathbf{b}_i\mathbf{a}_i^\top,
\end{equation}
where  $\Delta W_{(1,i)}=\mathbf{b}_i\mathbf{a}_i^\top$ is a rank-1 matrix formed by vectors $\mathbf{b}_i\in\mathbb{R}^{d\times 1}$ and $\mathbf{a}_i^\top\in\mathbb{R}^{1\times d}$. We treat this set of $r$ rank-1 matrices, $\{\Delta W_{(1,i)}\}_{i=1}^r$, as our \textbf{Rank-1 Expert Pool}.

To dynamically compose an update from this pool, we employ a lightweight router guided by the input's semantics. Inspired by the special role of the \texttt{[CLS]} token in capturing global information in Transformers  \cite{liang2022not, wang2024cls}, we use this feature representation $\phi(\mathbf{x})$ as the router's input. The router, a linear layer $\mathbf{W}_{router}\in\mathbb{R}^{r\times d_{CLS}}$, produces expert selection scores $\mathbf{s}_n=\mathbf{W}_{router}\phi(\mathbf{x}_n)$ for each input $\mathbf{x}_n$. The output is then a composition of the original weights and the selected experts:
\begin{align}
y = \sigma\left(W_0(x)+\sum_{i = 1}^{r}\pi_i(\mathbf{W}_{router}\phi(x))\mathbf{b}_i\mathbf{a}_i^\top x\right),
\end{align}
where \(W_0\) is the original weight matrix and \(\sigma\) is an activation function. The term \(\mathbf{W}_{router}\phi(x)\) calculates scores that determine the contribution of the \(i\)-th expert, \(\mathbf{b}_i\mathbf{a}_i^\top\). These scores are also used for the subsequent sparsification of LoRA. Finally, $\pi_i$ is a gating function that leverages these scores to determine the selection of the $i$-th expert.
During training, we apply a two-stage composition to select the most relevant $R$ experts for the current batch. First, for each sample, we identify the top-$R$ experts based on its scores, denoted by the index set $\mathcal{S}_n = \textit{Top}(\mathbf{s}_n, R)$. Second, we aggregate these selections across the batch to obtain an expert vote count $\mathbf{v}$, and select the overall top-$R$ experts for activation based on $\mathbf{v}$ in current batch, resulting in the final set $\mathcal{S}_{\text{batch}}=\textit{Top}(\mathbf{v}, R)$. This ensures that only the most critical experts for the current data distribution are updated. We also maintain an activation frequency memory, $\mathbf{C}_l$, and for each sample $\mathbf{x}_n$, $\mathbf{C}_l(i)=\mathbf{C}_l(i)+1$ if $i\in\mathcal{S}_n$. This dynamic selection mechanism allows the model to form sparse subspaces for different datasets, as discussed in ablation studies. 

\subsection{Activation-Guided Orthogonal (AGO) Loss}
Previous work suggests that LoRA parameters encapsulate crucial model update directions  \cite{wang2023orthogonal, yang2024parameter}. Therefore, learning in a subspace orthogonal to those of previous tasks can alleviate catastrophic forgetting. A standard approach is to enforce orthogonality between the LoRA matrix $\mathbf{B}_t$ of the current task $t$ and the concatenated matrix of all past tasks, $\mathbf{B}_{past}=\text{Concat}(\mathbf{B}_1,\cdots,\mathbf{B}_{t-1})$. We formalize the orthogonality constraint as follows:
\begin{equation}
L_{orth}=\frac{1}{mn}\sum_{i = 1}^{m}\sum_{j = 1}^{n}\Big|\left(\mathbf{b}_{past}^i\right)^\top\mathbf{b}_t^j\Big|,
\end{equation}
where we define $\mathbf{B}_{past}=[\mathbf{b}_{past}^1,\mathbf{b}_{past}^2,\cdots,\mathbf{b}_{past}^m]$ and $\mathbf{B}_t=[\mathbf{b}_t^1,\mathbf{b}_t^2,\cdots,\mathbf{b}_t^n]$, and $m$ and $n$ are the number of column vectors (basis vectors) in $\mathbf{B}_{past}$ and $\mathbf{B}_t$, respectively.

However, this naive orthogonality suffers from two key issues. First, it can lead to ``parameter collision''  \cite{yang2024parameter}, where parameters still interfere with each other despite the overall subspace orthogonality. Second, applying a strong orthogonal loss can compromise the optimization of the primary classification task, harming final performance.

To resolve this, we propose an \textbf{Activation-Guided Orthogonal (AGO) loss} that synergizes with our Rank-1 Expert Pool. Instead of enforcing orthogonality on the entire dense LoRA subspace, our approach uses the expert activation frequencies $\mathbf{C}_l$ recorded during previous compositions. We identify the set of the top-$R$ most critical experts from past tasks, denoted as $\mathcal{S}_{freq} = Top(\mathbf{C}_l, R)$. We then construct a sparse LoRA matrix, $\Delta W_{AGO}$, where only these  critical experts are retained for the loss calculation:
\begin{equation}
\Delta W_{AGO}=\sum_{i\in\mathcal{S}_{freq}}\mathbf{b}_i\mathbf{a}_i^\top + \sum_{i\notin\mathcal{S}_{freq}}\mathbf{0}.
\end{equation}
This means that ranks deemed unimportant for the current task, based on historical activation, are temporarily discarded here. By applying $L_{orth}$ only to these sparse, task-critical subspaces, we can effectively reduce parameter collision and isolate optimization directions without overly impacting the learning of the downstream classification task. 
\subsection{Training and Inference}
Our training objective combines a standard supervised classification loss, $L_{sup}$ (using the alignment loss from CLIP), with our proposed AGO loss, $L_{orth}$, described in previous subsection. The final loss is:
\begin{equation}
L = L_{sup}+\lambda L_{orth},
\end{equation}
where the hyper-parameter $\lambda$ balances the two terms.

Upon completion of training for each task, the learned LoRA weights are merged back into the main model's original weights, resulting in zero additional inference overhead. The decomposition and composition of the expert pool allows for flexible merging strategies at inference time. By default, to strike a balance between performance on the new task and retention of past knowledge, we merge only the top-$R$ / 2 experts as determined by their activation frequency. However, in scenarios where task-id is known  \cite{yu2024boosting}, our method can adaptively merge a more complete set of experts specific to that task, offering flexibility for practical situations. \textit{We provide a detailed analysis of our method's computational advantages in \textbf{Computation Cost} section.}

\setlength{\tabcolsep}{4.4pt}
\begin{table*}[t]

\small

    \begin{tabular}{lccccccccccccc}
            \hline 
        \textbf{Method} &\rotatebox{60}{\textbf{Task-ID}} & \rotatebox{60}{\textbf{Aircraft}} & \rotatebox{60}{\textbf{Caltech101}} & \rotatebox{60}{\textbf{CIFAR100}} & \rotatebox{60}{\textbf{DTD}} & \rotatebox{60}{\textbf{EuroSAT}} & \rotatebox{60}{\textbf{Flowers}} & \rotatebox{60}{\textbf{Food}} & \rotatebox{60}{\textbf{MNIST}} & \rotatebox{60}{\textbf{Pets}} & \rotatebox{60}{\textbf{Cars}} & \rotatebox{60}{\textbf{SUN397}} & \rotatebox{60}{\textbf{Average}} \\
        \midrule
        \multicolumn{12}{l}{\textbf{CLIP}} \\
        Zero-shot& \multirow{2}{*}{\rotatebox{90}{\textbf{TIF}}} & 24.3 & 88.4 & 68.2 & 44.6 & 54.9 & 71.0 & 88.5 & 59.6 & 89.0 & 64.7 & 65.2 & 65.3 \\
        5-shot Full Fine-tune & & 30.6 & 93.5 & 76.8 & 65.1 & 91.7 & 92.9 & 83.3 & 96.6 & 84.9 & 65.4 & 71.3 & 77.5 \\
        \midrule   
        \multicolumn{12}{l}{\textbf{Transfer}} \\
        Continual-FT & \multirow{7}{*}{\rotatebox{90}{\textbf{TIF}}} &-& 72.8 & 53.0 & 36.4 & 35.4 & 43.3 & 68.4 & 47.4 & 72.6 & 30.0 & 52.7 & 51.2   \\      
        LwF-VR \small{ {\texttt{[Arxiv'22]}}} &  &-& 82.2 & 62.5 & 40.1 & 40.1 & 56.3 & 80.0 & 60.9 & 77.6 & 40.5 &60.8 & 60.1  \\
        WiSE-FT \small{ {\texttt{[CVPR'22]}}} &  &-& 77.6 & 60.0 & 41.3 & 39.4 & 53.0 & 76.6 & 58.1 & 75.5 & 37.3 & 58.2 & 57.7  \\
        ZSCL \small{ {\texttt{[ICCV'23]}}}& &-& 84.0 & 68.1 & 44.8 & 46.8 & 63.6 & 84.9 & 61.4 & 81.4 & 55.5 & 62.2 & 65.3  \\
        GIFT \small{ {\texttt{[CVPR'25]}}}&  &-& \textbf{88.6} & 65.0 & \textbf{45.9} & 49.8 & 64.1 & 84.9 & \textbf{61.7} & 88.5 & 55.5 & \textbf{67.5} & 67.1\\
        \rowcolor{gray!30} 
        Ours  &  &-& 88.3 & \textbf{68.8} & 45.4 & \textbf{58.6} & \textbf{71.3} & \textbf{87.7} & 61.0 & \textbf{90.0} & \textbf{64.1} & 67.4 & \textbf{70.3} \small{(+3.2)} \\
        \midrule

        MoE-a \small{ {\texttt{[CVPR'24]}}}& \multirow{2}{*}{\rotatebox{90}{ \small{\textbf{TIK}}}} &-& 87.9 & 68.2 & 44.1 & 48.1 & 64.7 & \textbf{88.8} & \textbf{69.0} & 89.1 & 64.5 & 65.1 & 68.9 \\
        RAIL \small{ {\texttt{[NeurIPS'24]}}}&  &-& \textbf{88.4} & 68.2 & 44.6 & 54.9 & 71.0 & 88.5& 59.6 & 89.0 & \textbf{64.7} &65.2 & 69.4 \\
        \rowcolor{gray!30} 
        Ours  &  &-& 88.3 & \textbf{68.8} & \textbf{45.4} & \textbf{58.6} & \textbf{71.3} & 87.7 & 61.0 & \textbf{90.0} & 64.1 & \textbf{67.4} &\textbf{70.3} \small{(+0.9)} \\
        
         \midrule
    \multicolumn{12}{l}{\textbf{Average}} \\
    Continual-FT & \multirow{7}{*}{\rotatebox{90}{\textbf{TIF}}} & 28.1 & 86.4 & 59.1 & 52.8 & 55.8 & 62.0 & 70.2 & 64.7 & 75.5 & 35.0 & 54.0 & 58.5 \\
    LwF-VR \small{ {\texttt{[Arxiv'22]}}} &  & 24.9 & 89.1 & 64.2 & 53.4 & 54.3 & 70.8 & 79.2 & 66.5 & 79.2 & 44.1 & 61.6 & 62.5 \\
    WiSE-FT \small{ {\texttt{[CVPR'22]}}} &  & 32.0 & 87.7 & 61.0 & 55.8 & 68.1 & 69.3 & 76.8 & 71.5 & 77.6 & 42.0 & 59.3 & 63.7 \\
    ZSCL \small{ {\texttt{[ICCV'23]}}} &  & 28.2 & 88.6 & 66.5 & 53.5 & 56.3 & 73.4 & 83.1 & 56.4 & 82.4 & 57.5 & 62.9 & 64.4 \\
    GIFT \small{ {\texttt{[CVPR'25]}}} &  & 29.2 & 90.4 & 65.7 & 57.6 & 69.9 & 78.0 & 84.8 & 73.0 & 88.2 & 57.8 & \textbf{67.9} &69.3\\
    \rowcolor{gray!30} 
    Ours  &  & \textbf{33.4} & \textbf{91.4} & \textbf{74.5} & \textbf{59.6} & \textbf{77.8} & \textbf{80.0} & \textbf{88.0} & \textbf{73.4} & \textbf{90.9} & \textbf{65.4} & \textbf{67.9} & \textbf{73.0}  \small{(+3.7)}\\
    \midrule
    
    MoE-a \small{ {\texttt{[CVPR'24]}}} & \multirow{2}{*}{\rotatebox{90}{ \small{\textbf{TIK}}}} & 30.0 & 89.6 & 73.9 & 58.7 & 69.3 & 79.3 &88.1& \textbf{76.5} & 89.1 & 65.3 & 65.8 & 71.4 \\
    RAIL \small{ {\texttt{[NeurIPS'24]}}} &  & 32.9 & \textbf{94.5} & 69.9 & 58.1 & 71.8 & \textbf{84.4} & \textbf{88.5} & 70.4 & 89.0 & \textbf{66.1} & 65.7 & 71.9 \\
    \rowcolor{gray!30} 
    Ours  &  & \textbf{36.5} & 90.7 & \textbf{76.3} & \textbf{59.9} & \textbf{78.6} & 80.6 & 88.0 & 73.4 & \textbf{91.0} & 65.4 & \textbf{67.9}&\textbf{73.5}  \small{(+1.6)} \\
    
 \midrule

    \multicolumn{12}{l}{\textbf{Last}} \\
    Continual-FT & \multirow{7}{*}{\rotatebox{90}{\textbf{TIF}}} & 27.8 & 86.9 & 60.1 & 58.4 & 56.6 & 75.7 & 73.8 & 93.1 & 82.5 & 57.0 & 66.8 & 67.1 \\
    LwF-VR \small{ {\texttt{[Arxiv'22]}}} &  & 22.9 & 89.8 & 59.3 & 57.1 & 57.6 & 79.2 & 78.3 & 77.7 & 83.6 & 60.1 & 69.8 & 66.9 \\
    WiSE-FT \small{ {\texttt{[CVPR'22]}}} &  & 30.8 & 88.9 & 59.6 & 60.3 & 80.9 & 81.7 & 77.1 & 94.9 & 83.2 & 62.8 & 70.0 & 71.9 \\
    ZSCL \small{ {\texttt{[ICCV'23]}}} &  & 26.8 & 88.5 & 63.7 & 55.7 & 60.2 & 82.1 & 82.6 & 58.6 & 85.9 & 66.7 & 70.4 & 67.4 \\
    GIFT \small{ {\texttt{[CVPR'25]}}} &  & 27.9 & 89.8 & 46.0 & 62.0 & 71.9 & \textbf{87.8} & 83.4 & 93.0 & 86.1 & 67.6 & 71.8&71.6\\
    \rowcolor{gray!30} 
    Ours &  & \textbf{31.1} & \textbf{93.0} & \textbf{75.0} & \textbf{64.1} & \textbf{87.4} & 85.4 & \textbf{88.5} & \textbf{95.1} & \textbf{93.5} & \textbf{71.2} & \textbf{72.8} & \textbf{77.9}  \small{(+6.0)} \\
    \midrule
    MoE-a \small{ {\texttt{[CVPR'24]}}} & \multirow{2}{*}{\rotatebox{90}{ \small{\textbf{TIK}}}} & 30.1 & 89.3 & 74.9 & 64.0 & 82.3 & 89.4 & 87.1 & 89.0 & 89.1 & 69.5 & 72.5 & 76.1 \\
    RAIL \small{ {\texttt{[NeurIPS'24]}}} &  & 32.9 & \textbf{95.1} & 70.3 & 63.2 & 81.5 & \textbf{95.6} & \textbf{88.5} & 89.7 & 89.0 & \textbf{72.5} & 71.0 & 77.2 \\
    \rowcolor{gray!30} 
    Ours   &  & \textbf{36.5} & 90.9 & \textbf{77.9} & \textbf{65.3} & \textbf{90.1} & 88.4 & 88.4 & \textbf{95.2} & \textbf{93.7} & 71.4 & \textbf{72.8}& \textbf{79.2}  \small{(+2.0)} \\
    
        \midrule
    \end{tabular}
        \caption{Comparisons with state-of-the-art methods on few-shot MTIL Order 1 benchmark in terms of ``Transfer'', ``Average'', and ``Last'' scores (\%). \textbf{TIK} means Task-Id Known, and \textbf{TIF} means Task-Id Free. We label the best method with bold style.}
        \label{tab:comparison1}
\end{table*}

\section{Experiments}
\subsection{Experimental Setting}
In this part, we detail the experimental setup employed to evaluate our proposed method. Our experiments mainly focus on classic and challenging learning scenarios for CLIP.

\textbf{Datasets} \quad In the Multi-domain Task Incremental Learning (MTIL) setting  and the Task-Agnostic Incremental Learning (X-TAIL) setting, we utilize a total of 11 datasets: Aircraft  \cite{maji2013fine}, Caltech101  \cite{fei2004learning}, CIFAR 100  \cite{krizhevsky2009learning}, DTD  \cite{cimpoi2014describing}, EuroSAT  \cite{helber2019eurosat}, Flowers  \cite{nilsback2008automated}, Food  \cite{bossard2014food}, MNIST  \cite{deng2012mnist}, OxfordPet  \cite{parkhi2012cats}, StanfordCars  \cite{krause20133d}, and SUN397  \cite{xiao2010sun}. Each dataset is treated as an individual task for continual learning. Specifically, we adopt a 5-shot split for both MTIL and X-TAIL  \cite{yu2024boosting,xu2024advancing}.

\textbf{Metrics} \quad Following previous works  \cite{zheng2023preventing}, we use three standard metrics: ``Transfer'' (generalization to unseen data), ``Average'' (overall performance across all tasks), and ``Last'' (knowledge retention from past tasks).

\textbf{Implementation Details} \quad We follow the overall setups in  \cite{zheng2023preventing,yu2024boosting}, using a CLIP ViT-B/16 backbone with our technique applied to all linear layers. Key hyperparameters are: LoRA rank set to 12, with 8 ranks retained after sparsification and the top 4 experts merged after training. The orthogonal loss weight $\lambda$ is 0.1. We train for 500 iterations per task using the AdamW optimizer with a batch size of 32 and a learning rate of 2e-3. All experiments were conducted on RTX A6000 GPUs.

\subsection{Comparisons with State-Of-The-Arts}

We compare our method with traditional continual learning baselines (LWF-VR  \cite{ding2022don}, WISE-FT  \cite{wortsman2022robust}) and recent VLM-specific methods (ZSCL  \cite{zheng2023preventing}, MoE-a  \cite{yu2024boosting}, RAIL  \cite{xu2024advancing}, GIFT  \cite{wu2025synthetic}).

As presented in Table \ref{tab:comparison1}, here 65.3\% is the CLIP's average across all 11 tasks. It can be seen that our method achieves the best performance regardless of whether task-id prior knowledge is available or not. Specifically, when task-id is not given, our method outperforms the previous state-of-the-art (SOTA) methods by 3.2\%, 3.7\%, 6.0\% in the Transfer, Average, Last metric, indicating that the method not only improves the performance of downstream tasks, but also better preserves the ability of pre-trained weights. It is worth noting that here our method even does not introduce any additional external large-scale datasets or synthetic data replay. Moreover, when task-id is given, our method still has an edge over previous methods, with improvements of 0.9\%, 1.6\%, and 2.0\% respectively in the corresponding metrics. Notably, our method exceeds the zero-shot upper bound performance of CLIP in the Transfer metric, and consistently shows a steady improvement over all datasets during continual learning, which demonstrates the excellent generalization ability of our method. 

\setlength{\tabcolsep}{5pt}
\begin{table}[h]
\small

  \begin{tabular}{llcl}
    \toprule
   \textbf{Method} & \makecell{Training\\ Param. (MB)} & \makecell{Extra Data \& \\Components} & \makecell{GPU Mem\\ Cost (MB)} \\
    \midrule
    ZSCL & 570.76 \small{ {(-0)}}& IN & 28454\small{ {(-0)}}\\
    MoE-a & 194.78 \underline{\small{ {(-65.5\%)}}} & AN \& TN & 14040 \underline{\small{ {(-50.7\%)}}}\\
    GIFT & 570.76  \small{ {(-0)}} & SD & 22990 \underline{\small{ {(-19.2\%)}}}\\
    \rowcolor{gray!30}
    Ours & \textbf{18.99}  \underline{\small{ {(-96.7\%)}}} &-& \textbf{9490} \underline{\small{ {(-66.7\%)}}}\\
    
    \bottomrule
  \end{tabular}
    \caption{Computational Cost Comparison (IN: ImageNet1K, AN: AlexNet, TN: TinyImageNet, SD: Stable Diffusion) .} 
  \label{tab:comp_cost} 

\end{table}

\subsection{Computation Cost}
As shown in Table \ref{tab:comp_cost}, our framework is remarkably computationally efficient. Compared to full fine-tuning baselines  \cite{zheng2023preventing, wu2025synthetic,yu2024boosting}, our approach reduces trainable parameters by \textbf{96.7\%} and peak GPU memory consumption by \textbf{66.7\%}. Moreover, Our method's training/inference speeds are 1.84/2.98 it/s, compared to MoE-a's 0.78/1.04 it/s for our method's  adapter-free style, introducing zero inference overhead. Crucially, these efficiency gains are achieved without requiring any extra datasets or additional components (data replay techniques \cite{zheng2023preventing} or additional adapters \cite{xu2024advancing}), which have been commonly used in these previous works, showing our method's practicality for realistic continual learning scenarios.

\section{Discussion}

\subsection{Analysis of Method Components}

We ablate the effects of our method's key components in Table \ref{tab:ablation}. Both the dynamic composition from our Rank-1 Expert Pool and the \textbf{Activation-Guided Orthogonal (AGO) loss} individually contribute positively to performance when compared to baselines. Specifically, the dynamic composition of rank-1 experts prevents redundant weight updates, enhancing transfer ability and improving overall classification performance. The AGO loss, in turn, focuses the model on critical parameter updates, boosting downstream task accuracy. When the two are combined, they achieve the best results, demonstrating a synergistic effect.

Notably, all metrics in this section are presented as (\%), and all ablation studies are conducted on the same experimental setting as Table \ref{tab:comparison1}, unless otherwise stated.

\setlength{\tabcolsep}{5.5pt}
\begin{table}[t]
\small
  \begin{tabular}{lcccccc}
    \toprule
    Method & Trans. & $\Delta$ & Avg. & $\Delta$ & Last & $\Delta$ \\
    \midrule
    \textbf{Baselines} & \multicolumn{5}{c}{} \\ 
    Zero-shot & 69.4 & 0.0 & 65.3 & 0.0 & 65.3 & 0.0 \\
    ZSCL & 65.3 &   {-4.1} & 64.4 &   {-0.9} & 67.4 &  \underline{+2.1} \\
    SD-LoRA & 67.4 &   {-2.0} & 70.1 &  \underline{+4.8} & 74.2 &  \underline{+8.9} \\
    LoRAMoE & 68.0 &   {-1.4} & 69.9 &  \underline{+4.6} & 70.3 &  \underline{+5.0} \\
    \midrule
    \textbf{Ablation} & \multicolumn{5}{c}{} \\ 
    Vanilla LoRA & 65.0 &   {-4.4} & 68.5 &  \underline{+3.2} & 73.8 &  \underline{+8.5} \\
    + Expert Pool & 69.7 &  \underline{+0.3} & 72.4 &  \underline{+7.1} & 77.1 &  \underline{+11.8} \\
    + AGO loss & 65.3 &   {-4.1} & 69.2 &  \underline{+3.9} & 74.5 &  \underline{+9.2} \\
    \rowcolor{gray!30}
    +Ours & \textbf{70.3} &  \underline{+0.9} & \textbf{73.0} &  \underline{+7.7} & \textbf{77.9} &  \underline{+12.6} \\
    \bottomrule
  \end{tabular}
    \caption{Ablation study on method components, conducted on MTIL setting Order 1. Underline means improvement.} 
  \label{tab:ablation}
    
\end{table}

\begin{figure}[t]
  \centering
  
  \includegraphics[width=1\linewidth]{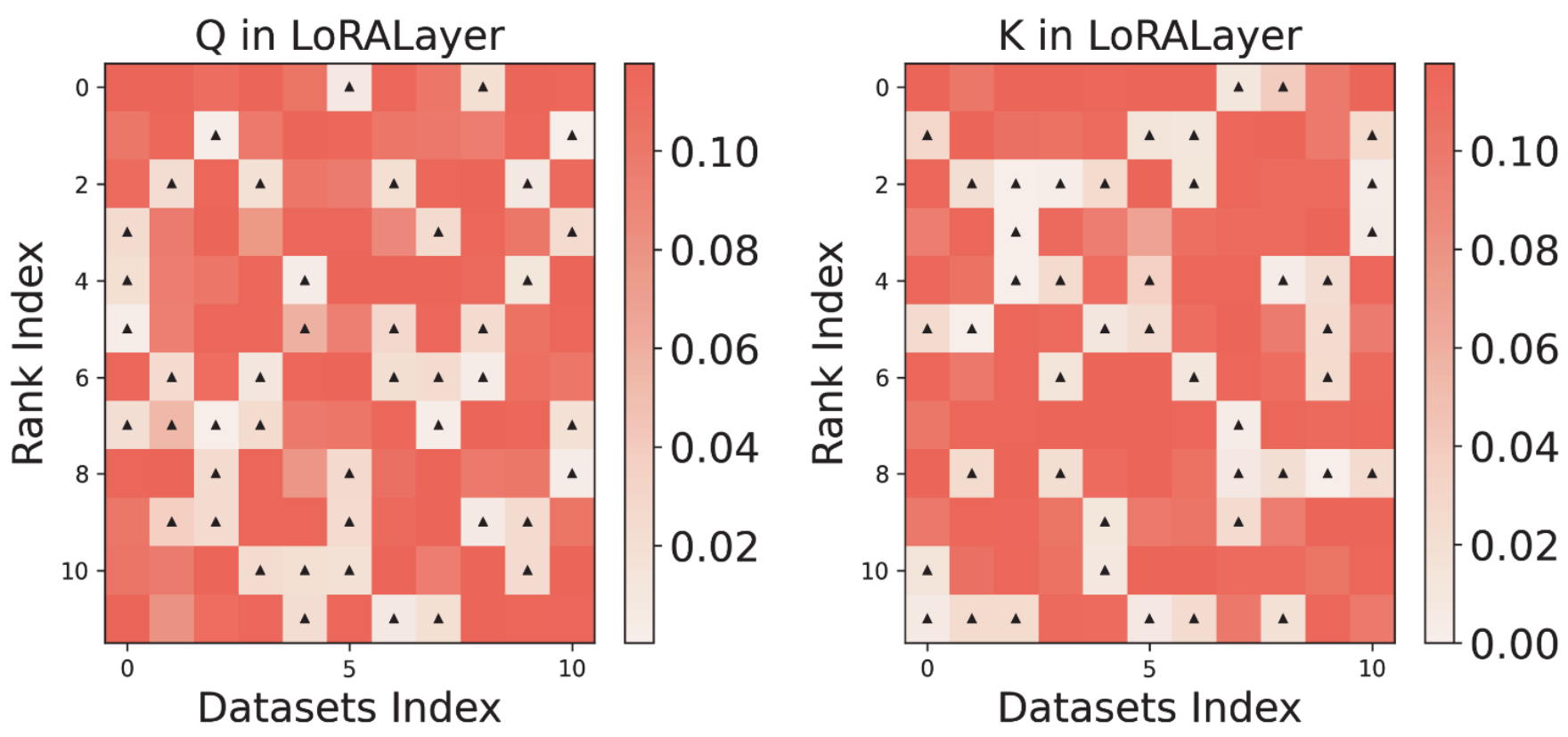}
  \caption{ Visualization of the activation frequencies of LoRA under different tasks. The darker the color, the higher frequency. ``$\blacktriangle$'' means disregarded experts. This figure shows frequency of the first LoRA layer in image encoder, indicating that the composed part of each task is exactly the experts with relatively higher activation. }
  
  \label{fig:act}
  
\end{figure}
\subsection{Discussion on LoRA-based CL method}

We also present other LoRA-based continual learning methods in Table \ref{tab:ablation}: SD-LoRA  \cite{wu2025sdlora} and LoRAMoE  \cite{dou2023loramoe}, which are primarily designed to address simple static incremental problems. As shown, their performance is suboptimal when facing fine-grained dynamic domain discrepancies, where our approach performs better.

\subsection{Analysis of Expert Composition}

To verify that our framework selects domain-specific experts from the expert pool, we visualize the expert activation frequencies in Figure \ref{fig:act}. The heatmaps show distinct activation patterns across different datasets (columns), confirming that the \texttt{[CLS]} token effectively guides the router to compose domain-aware subspaces. 

We also study the impact of which experts are chosen for the final merged LoRA in Table \ref{tab:moe-sparsity}. The results clearly indicate that merging the experts with the highest activation frequency (``Top'') yields the best performance. Conversely, using low-frequency experts (``Down'') critically degrades performance. Interestingly, a highly sparse composition (Top-2) further improves the Transfer metric, likely due to extreme fewer parameter updates enhancing generalization. Our default setting (Top-4) is chosen to strike an optimal balance between downstream performance and generalization.

\setlength{\tabcolsep}{5.5pt}
\begin{table}[t]

\small
  \begin{tabular}{lcccccc}
    \toprule
    \textbf{Strategy} & Trans. & $\Delta$ & Avg. & $\Delta$ & Last & $\Delta$ \\
    \midrule
    \multicolumn{7}{l}{\textbf{Great Experts}} \\
    Top-2 (2 E) & 70.5 &  \underline{+0.2} & 71.5 &   {-1.5} & 75.3 &   {-2.6} \\
    \rowcolor{gray!30}
    \textbf{Top-4 (Ours)} & \textbf{70.3} & 0.0 & \textbf{73.0} & 0.0 & \textbf{77.9} & 0.0 \\
    Top-8 (8 E) & 67.2 &   {-3.1} & 69.6 &   {-3.4} & 73.7 &   {-4.2} \\
    
    \midrule
    \multicolumn{7}{l}{\textbf{Good Experts}} \\
    Mid-4 (4 E) & 70.3 & 0.0 & 72.8 &   {-0.2} & 77.4 &   {-0.5} \\
    
    \midrule
    \multicolumn{7}{l}{\textbf{Bad Experts}} \\
    Down-4 (4 E) & 68.4 &   {-1.9} & 64.0 &   {-9.0} & 62.9 &   {-15.0} \\
    Down-8 (8 E) & 68.3 &   {-2.0} & 70.2 &   {-2.8} & 72.4 &   {-5.5} \\
    \bottomrule
    
  \end{tabular}
  
  \caption{Ablation study on rank composing selection, involving expert number and frequency selection, 2 E means selecting 2 experts for sparsification. }
  \label{tab:moe-sparsity}
  
\end{table}

\begin{figure}[t]
  \centering
  
  \includegraphics[width=1\linewidth]{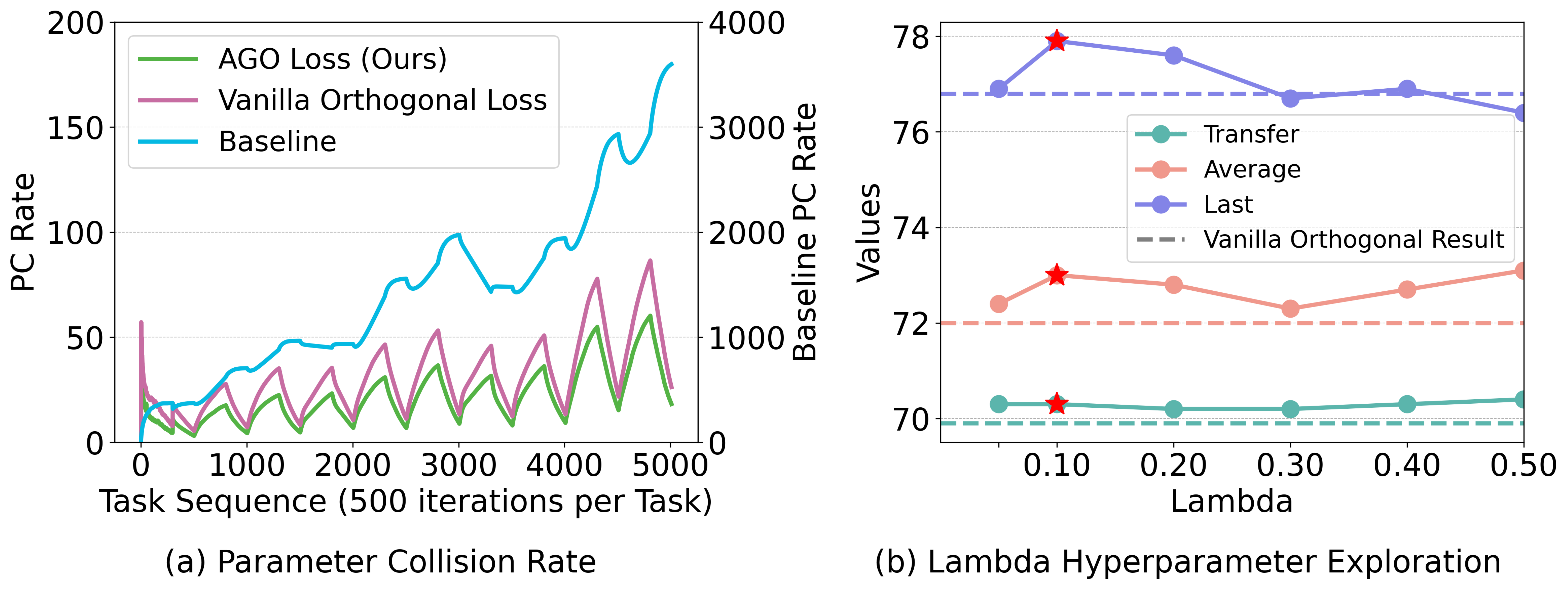}
  \caption{(a) Parameter collision rate comparison: Our method has fewer parameter collisions than vanilla orthogonal LoRA and non-orthogonal LoRA. (b) $\lambda$ hyperparameter exploration: Our method is robust to $\lambda$, outperforming vanilla orthogonal LoRA in all metrics.}
  
  \label{fig:sol}
  
\end{figure}

\subsection{Analysis of Activation-Guided Orthogonal Loss}

We analyze the advantage of AGO loss by measuring the parameter collision rate, where a higher $L_{orth}$ value implies more collisions. As shown in Figure \ref{fig:sol}(a), our activation-guided approach maintains a  lower collision rate than both a non-orthogonal baseline and a vanilla (dense) orthogonal loss, effectively isolating task-specific updates.
Figure \ref{fig:sol}(b) explores the impact of the loss weight, $\lambda$, which defines the proportion of our AGO loss. The figure shows that as the proportion of the orthogonal loss increases, downstream task performance is affected while the generalization ability (Transfer) rises. The results also show our method is robust to the choice of $\lambda$, with $\lambda=0.1$ providing the best balance.

\section{Conclusion}
In this paper, we introduced an efficient continual learning framework by restructuring a single LoRA module into a \textbf{decomposable} Rank-1 Expert Pool. Guided by a lightweight, semantics-aware router, our method dynamically \textbf{composes} sparse, task-specific subspaces from this pool. These composed subspaces are further decoupled by a synergistic \textbf{Activation-Guided Orthogonal (AGO) loss}. Extensive experiments demonstrate that our approach achieves state-of-the-art results across various continual learning settings while eliminating additional inference costs.

\section{Acknowledgments}
This work is supported by NSFC Project (62222604, 62192783, 624B2063, 62506162, 62476136), Jiangsu Science and Technology Project (BK20251241, BK20250142).

\bibliography{aaai2026}

@String{Computing = "Computing" }

@String{Computer = "{IEEE} Computer" }

@String{Springer = "Springer-Verlag" }

@ArtifactSoftware{R,
    title = {R: A Language and Environment for Statistical Computing},
    author = {{R Core Team}},
    organization = {R Foundation for Statistical Computing},
    address = {Vienna, Austria},
    year = {2019},
    url = {https://www.R-project.org/},
}

@article{liang2022not,
  title={Not all patches are what you need: Expediting vision transformers via token reorganizations},
  author={Liang, Youwei and Ge, Chongjian and Tong, Zhan and Song, Yibing and Wang, Jue and Xie, Pengtao},
  journal={arXiv preprint arXiv:2202.07800},
  year={2022}
}

@inproceedings{rebuffi2017icarl,
  title={icarl: Incremental classifier and representation learning},
  author={Rebuffi, Sylvestre-Alvise and Kolesnikov, Alexander and Sperl, Georg and Lampert, Christoph H},
  booktitle={Proceedings of the IEEE conference on Computer Vision and Pattern Recognition},
  pages={2001--2010},
  year={2017}
}

@article{jha2024clap4clip,
  title={Clap4clip: Continual learning with probabilistic finetuning for vision-language models},
  author={Jha, Saurav and Gong, Dong and Yao, Lina},
  journal={arXiv preprint arXiv:2403.19137},
  year={2024}
}

@inproceedings{yu2024boosting,
  title={Boosting continual learning of vision-language models via mixture-of-experts adapters},
  author={Yu, Jiazuo and Zhuge, Yunzhi and Zhang, Lu and Hu, Ping and Wang, Dong and Lu, Huchuan and He, You},
  booktitle={Proceedings of the IEEE/CVF Conference on Computer Vision and Pattern Recognition},
  pages={23219--23230},
  year={2024}
}

@inproceedings{zheng2023preventing,
  title={Preventing zero-shot transfer degradation in continual learning of vision-language models},
  author={Zheng, Zangwei and Ma, Mingyuan and Wang, Kai and Qin, Ziheng and Yue, Xiangyu and You, Yang},
  booktitle={Proceedings of the IEEE/CVF international conference on computer vision},
  pages={19125--19136},
  year={2023}
}

@article{thengane2022clip,
  title={Clip model is an efficient continual learner},
  author={Thengane, Vishal and Khan, Salman and Hayat, Munawar and Khan, Fahad},
  journal={arXiv preprint arXiv:2210.03114},
  year={2022}
}

@article{gao2024higher,
  title={Higher layers need more lora experts},
  author={Gao, Chongyang and Chen, Kezhen and Rao, Jinmeng and Sun, Baochen and Liu, Ruibo and Peng, Daiyi and Zhang, Yawen and Guo, Xiaoyuan and Yang, Jie and Subrahmanian, VS},
  journal={arXiv preprint arXiv:2402.08562},
  year={2024}
}

@article{dou2023loramoe,
  title={LoRAMoE: Alleviate world knowledge forgetting in large language models via MoE-style plugin},
  author={Dou, Shihan and Zhou, Enyu and Liu, Yan and Gao, Songyang and Zhao, Jun and Shen, Wei and Zhou, Yuhao and Xi, Zhiheng and Wang, Xiao and Fan, Xiaoran and others},
  journal={arXiv preprint arXiv:2312.09979},
  year={2023}
}

@article{xu2024advancing,
  title={Advancing Cross-domain Discriminability in Continual Learning of Vision-Language Models},
  author={Xu, Yicheng and Chen, Yuxin and Nie, Jiahao and Wang, Yusong and Zhuang, Huiping and Okumura, Manabu},
  journal={arXiv preprint arXiv:2406.18868},
  year={2024}
}

@article{ding2023sparse,
  title={Sparse low-rank adaptation of pre-trained language models},
  author={Ding, Ning and Lv, Xingtai and Wang, Qiaosen and Chen, Yulin and Zhou, Bowen and Liu, Zhiyuan and Sun, Maosong},
  journal={arXiv preprint arXiv:2311.11696},
  year={2023}
}

@article{wang2023orthogonal,
  title={Orthogonal subspace learning for language model continual learning},
  author={Wang, Xiao and Chen, Tianze and Ge, Qiming and Xia, Han and Bao, Rong and Zheng, Rui and Zhang, Qi and Gui, Tao and Huang, Xuanjing},
  journal={arXiv preprint arXiv:2310.14152},
  year={2023}
}

@article{yang2024parameter,
  title={Is Parameter Collision Hindering Continual Learning in LLMs?},
  author={Yang, Shuo and Ning, Kun-Peng and Liu, Yu-Yang and Yao, Jia-Yu and Tian, Yong-Hong and Song, Yi-Bing and Yuan, Li},
  journal={arXiv preprint arXiv:2410.10179},
  year={2024}
}

@article{mao2024dora,
  title={Dora: Enhancing parameter-efficient fine-tuning with dynamic rank distribution},
  author={Mao, Yulong and Huang, Kaiyu and Guan, Changhao and Bao, Ganglin and Mo, Fengran and Xu, Jinan},
  journal={arXiv preprint arXiv:2405.17357},
  year={2024}
}

@article{feng2025omoe,
  title={OMoE: Diversifying Mixture of Low-Rank Adaptation by Orthogonal Finetuning},
  author={Feng, Jinyuan and Pu, Zhiqiang and Hu, Tianyi and Li, Dongmin and Ai, Xiaolin and Wang, Huimu},
  journal={arXiv preprint arXiv:2501.10062},
  year={2025}
}

@article{wu2025synthetic,
  title={Synthetic Data is an Elegant GIFT for Continual Vision-Language Models},
  author={Wu, Bin and Shi, Wuxuan and Wang, Jinqiao and Ye, Mang},
  journal={arXiv preprint arXiv:2503.04229},
  year={2025}
}

@article{wang2024cls,
  title={[CLS] Token Tells Everything Needed for Training-free Efficient MLLMs},
  author={Wang, Ao and Sun, Fengyuan and Chen, Hui and Lin, Zijia and Han, Jungong and Ding, Guiguang},
  journal={arXiv preprint arXiv:2412.05819},
  year={2024}
}

@inproceedings{radford2021learning,
  title={Learning transferable visual models from natural language supervision},
  author={Radford, Alec and Kim, Jong Wook and Hallacy, Chris and Ramesh, Aditya and Goh, Gabriel and Agarwal, Sandhini and Sastry, Girish and Askell, Amanda and Mishkin, Pamela and Clark, Jack and others},
  booktitle={International conference on machine learning},
  pages={8748--8763},
  year={2021},
  organization={PmLR}
}

@article{li2017learning,
  title={Learning without forgetting},
  author={Li, Zhizhong and Hoiem, Derek},
  journal={IEEE transactions on pattern analysis and machine intelligence},
  volume={40},
  number={12},
  pages={2935--2947},
  year={2017},
  publisher={IEEE}
}

@inproceedings{deng2009imagenet,
  title={Imagenet: A large-scale hierarchical image database},
  author={Deng, Jia and Dong, Wei and Socher, Richard and Li, Li-Jia and Li, Kai and Fei-Fei, Li},
  booktitle={2009 IEEE conference on computer vision and pattern recognition},
  pages={248--255},
  year={2009},
  organization={Ieee}
}

@article{hu2022lora,
  title={Lora: Low-rank adaptation of large language models.},
  author={Hu, Edward J and Shen, Yelong and Wallis, Phillip and Allen-Zhu, Zeyuan and Li, Yuanzhi and Wang, Shean and Wang, Lu and Chen, Weizhu and others},
  journal={ICLR},
  volume={1},
  number={2},
  pages={3},
  year={2022}
}

@article{zhang2023increlora,
  title={Increlora: Incremental parameter allocation method for parameter-efficient fine-tuning},
  author={Zhang, Feiyu and Li, Liangzhi and Chen, Junhao and Jiang, Zhouqiang and Wang, Bowen and Qian, Yiming},
  journal={arXiv preprint arXiv:2308.12043},
  year={2023}
}

@article{valipour2022dylora,
  title={Dylora: Parameter efficient tuning of pre-trained models using dynamic search-free low-rank adaptation},
  author={Valipour, Mojtaba and Rezagholizadeh, Mehdi and Kobyzev, Ivan and Ghodsi, Ali},
  journal={arXiv preprint arXiv:2210.07558},
  year={2022}
}

@article{meng2024pissa,
  title={Pissa: Principal singular values and singular vectors adaptation of large language models},
  author={Meng, Fanxu and Wang, Zhaohui and Zhang, Muhan},
  journal={Advances in Neural Information Processing Systems},
  volume={37},
  pages={121038--121072},
  year={2024}
}

@inproceedings{wortsman2022robust,
  title={Robust fine-tuning of zero-shot models},
  author={Wortsman, Mitchell and Ilharco, Gabriel and Kim, Jong Wook and Li, Mike and Kornblith, Simon and Roelofs, Rebecca and Lopes, Raphael Gontijo and Hajishirzi, Hannaneh and Farhadi, Ali and Namkoong, Hongseok and others},
  booktitle={Proceedings of the IEEE/CVF conference on computer vision and pattern recognition},
  pages={7959--7971},
  year={2022}
}

@article{ding2022don,
  title={Don't stop learning: Towards continual learning for the clip model},
  author={Ding, Yuxuan and Liu, Lingqiao and Tian, Chunna and Yang, Jingyuan and Ding, Haoxuan},
  journal={arXiv preprint arXiv:2207.09248},
  year={2022}
}

@article{zhang2023adalora,
  title={Adalora: Adaptive budget allocation for parameter-efficient fine-tuning},
  author={Zhang, Qingru and Chen, Minshuo and Bukharin, Alexander and Karampatziakis, Nikos and He, Pengcheng and Cheng, Yu and Chen, Weizhu and Zhao, Tuo},
  journal={arXiv preprint arXiv:2303.10512},
  year={2023}
}

@inproceedings{liu2024moe,
  title={When moe meets llms: Parameter efficient fine-tuning for multi-task medical applications},
  author={Liu, Qidong and Wu, Xian and Zhao, Xiangyu and Zhu, Yuanshao and Xu, Derong and Tian, Feng and Zheng, Yefeng},
  booktitle={Proceedings of the 47th International ACM SIGIR Conference on Research and Development in Information Retrieval},
  pages={1104--1114},
  year={2024}
}

@article{yang2024corda,
  title={CorDA: Context-Oriented Decomposition Adaptation of Large Language Models for Task-Aware Parameter-Efficient Fine-tuning},
  author={Yang, Yibo and Li, Xiaojie and Zhou, Zhongzhu and Song, Shuaiwen and Wu, Jianlong and Nie, Liqiang and Ghanem, Bernard},
  journal={Advances in Neural Information Processing Systems},
  volume={37},
  pages={71768--71791},
  year={2024}
}

@article{ren2024analyzing,
  title={Analyzing and reducing catastrophic forgetting in parameter efficient tuning},
  author={Ren, Weijieying and Li, Xinlong and Wang, Lei and Zhao, Tianxiang and Qin, Wei},
  journal={arXiv preprint arXiv:2402.18865},
  year={2024}
}

@inproceedings{agarwal2022semantics,
  title={Semantics-driven generative replay for few-shot class incremental learning},
  author={Agarwal, Aishwarya and Banerjee, Biplab and Cuzzolin, Fabio and Chaudhuri, Subhasis},
  booktitle={Proceedings of the 30th ACM international conference on multimedia},
  pages={5246--5254},
  year={2022}
}

@misc{rombach2021highresolution,
      title={High-Resolution Image Synthesis with Latent Diffusion Models}, 
      author={Robin Rombach and Andreas Blattmann and Dominik Lorenz and Patrick Esser and Björn Ommer},
      year={2021},
      eprint={2112.10752},
      archivePrefix={arXiv},
      primaryClass={cs.CV}
}

@inproceedings{jiang2025fine,
  title={Fine-tuning with Reserved Majority for Noise Reduction},
  author={Jiang, Shuyang and Liao, Yusheng and Zhang, Ya and Wang, Yanfeng and Wang, Yu},
  booktitle={The Thirteenth International Conference on Learning Representations},
  year={2025}
}

@article{wei2024online,
  title={Online-LoRA: Task-free Online Continual Learning via Low Rank Adaptation},
  author={Wei, Xiwen and Li, Guihong and Marculescu, Radu},
  journal={arXiv preprint arXiv:2411.05663},
  year={2024}
}

@article{gekhman2024does,
  title={Does fine-tuning LLMs on new knowledge encourage hallucinations?},
  author={Gekhman, Zorik and Yona, Gal and Aharoni, Roee and Eyal, Matan and Feder, Amir and Reichart, Roi and Herzig, Jonathan},
  journal={arXiv preprint arXiv:2405.05904},
  year={2024}
}

@inproceedings{parelli2023clip,
  title={Clip-guided vision-language pre-training for question answering in 3d scenes},
  author={Parelli, Maria and Delitzas, Alexandros and Hars, Nikolas and Vlassis, Georgios and Anagnostidis, Sotirios and Bachmann, Gregor and Hofmann, Thomas},
  booktitle={Proceedings of the IEEE/CVF Conference on Computer Vision and Pattern Recognition},
  pages={5607--5612},
  year={2023}
}

@inproceedings{antol2015vqa,
  title={Vqa: Visual question answering},
  author={Antol, Stanislaw and Agrawal, Aishwarya and Lu, Jiasen and Mitchell, Margaret and Batra, Dhruv and Zitnick, C Lawrence and Parikh, Devi},
  booktitle={Proceedings of the IEEE international conference on computer vision},
  pages={2425--2433},
  year={2015}
}

@inproceedings{hong2024navigating,
  title={Navigating Beyond Instructions: Vision-and-Language Navigation in Obstructed Environments},
  author={Hong, Haodong and Wang, Sen and Huang, Zi and Wu, Qi and Liu, Jiajun},
  booktitle={Proceedings of the 32nd ACM International Conference on Multimedia},
  pages={7639--7648},
  year={2024}
}

@article{maji2013fine,
  title={Fine-grained visual classification of aircraft},
  author={Maji, Subhransu and Rahtu, Esa and Kannala, Juho and Blaschko, Matthew and Vedaldi, Andrea},
  journal={arXiv preprint arXiv:1306.5151},
  year={2013}
}

@inproceedings{fei2004learning,
  title={Learning generative visual models from few training examples: An incremental bayesian approach tested on 101 object categories},
  author={Fei-Fei, Li and Fergus, Rob and Perona, Pietro},
  booktitle={2004 conference on computer vision and pattern recognition workshop},
  pages={178--178},
  year={2004},
  organization={IEEE}
}

@article{krizhevsky2009learning,
  title={Learning multiple layers of features from tiny images},
  author={Krizhevsky, Alex and Hinton, Geoffrey and others},
  year={2009},
  publisher={Toronto, ON, Canada}
}

@inproceedings{cimpoi2014describing,
  title={Describing textures in the wild},
  author={Cimpoi, Mircea and Maji, Subhransu and Kokkinos, Iasonas and Mohamed, Sammy and Vedaldi, Andrea},
  booktitle={Proceedings of the IEEE conference on computer vision and pattern recognition},
  pages={3606--3613},
  year={2014}
}

@article{helber2019eurosat,
  title={Eurosat: A novel dataset and deep learning benchmark for land use and land cover classification},
  author={Helber, Patrick and Bischke, Benjamin and Dengel, Andreas and Borth, Damian},
  journal={IEEE Journal of Selected Topics in Applied Earth Observations and Remote Sensing},
  volume={12},
  number={7},
  pages={2217--2226},
  year={2019},
  publisher={IEEE}
}

@inproceedings{nilsback2008automated,
  title={Automated flower classification over a large number of classes},
  author={Nilsback, Maria-Elena and Zisserman, Andrew},
  booktitle={2008 Sixth Indian conference on computer vision, graphics \& image processing},
  pages={722--729},
  year={2008},
  organization={IEEE}
}

@inproceedings{bossard2014food,
  title={Food-101--mining discriminative components with random forests},
  author={Bossard, Lukas and Guillaumin, Matthieu and Van Gool, Luc},
  booktitle={Computer vision--ECCV 2014: 13th European conference, zurich, Switzerland, September 6-12, 2014, proceedings, part VI 13},
  pages={446--461},
  year={2014},
  organization={Springer}
}

@article{deng2012mnist,
  title={The mnist database of handwritten digit images for machine learning research [best of the web]},
  author={Deng, Li},
  journal={IEEE signal processing magazine},
  volume={29},
  number={6},
  pages={141--142},
  year={2012},
  publisher={IEEE}
}

@inproceedings{parkhi2012cats,
  title={Cats and dogs},
  author={Parkhi, Omkar M and Vedaldi, Andrea and Zisserman, Andrew and Jawahar, CV},
  booktitle={2012 IEEE conference on computer vision and pattern recognition},
  pages={3498--3505},
  year={2012},
  organization={IEEE}
}

@inproceedings{krause20133d,
  title={3d object representations for fine-grained categorization},
  author={Krause, Jonathan and Stark, Michael and Deng, Jia and Fei-Fei, Li},
  booktitle={Proceedings of the IEEE international conference on computer vision workshops},
  pages={554--561},
  year={2013}
}

@inproceedings{xiao2010sun,
  title={Sun database: Large-scale scene recognition from abbey to zoo},
  author={Xiao, Jianxiong and Hays, James and Ehinger, Krista A and Oliva, Aude and Torralba, Antonio},
  booktitle={2010 IEEE computer society conference on computer vision and pattern recognition},
  pages={3485--3492},
  year={2010},
  organization={IEEE}
}

@article{lavda2018continual,
  title={Continual classification learning using generative models},
  author={Lavda, Frantzeska and Ramapuram, Jason and Gregorova, Magda and Kalousis, Alexandros},
  journal={arXiv preprint arXiv:1810.10612},
  year={2018}
}

@inproceedings{aljundi2018memory,
  title={Memory aware synapses: Learning what (not) to forget},
  author={Aljundi, Rahaf and Babiloni, Francesca and Elhoseiny, Mohamed and Rohrbach, Marcus and Tuytelaars, Tinne},
  booktitle={Proceedings of the European conference on computer vision (ECCV)},
  pages={139--154},
  year={2018}
}

@inproceedings{douillard2022dytox,
  title={Dytox: Transformers for continual learning with dynamic token expansion},
  author={Douillard, Arthur and Ram{\'e}, Alexandre and Couairon, Guillaume and Cord, Matthieu},
  booktitle={Proceedings of the IEEE/CVF conference on computer vision and pattern recognition},
  pages={9285--9295},
  year={2022}
}

@inproceedings{li2019learn,
  title={Learn to grow: A continual structure learning framework for overcoming catastrophic forgetting},
  author={Li, Xilai and Zhou, Yingbo and Wu, Tianfu and Socher, Richard and Xiong, Caiming},
  booktitle={International conference on machine learning},
  pages={3925--3934},
  year={2019},
  organization={PMLR}
}

@inproceedings{lee2023pre,
  title={Do pre-trained models benefit equally in continual learning?},
  author={Lee, Kuan-Ying and Zhong, Yuanyi and Wang, Yu-Xiong},
  booktitle={Proceedings of the IEEE/CVF Winter Conference on Applications of Computer Vision},
  pages={6485--6493},
  year={2023}
}

@inproceedings{
wu2025sdlora,
title={{SD}-Lo{RA}: Scalable Decoupled Low-Rank Adaptation for Class Incremental Learning},
author={Yichen Wu and Hongming Piao and Long-Kai Huang and Renzhen Wang and Wanhua Li and Hanspeter Pfister and Deyu Meng and Kede Ma and Ying Wei},
booktitle={The Thirteenth International Conference on Learning Representations},
year={2025},
url={https://openreview.net/forum?id=5U1rlpX68A}
}

\clearpage

\appendix

\section*{Appendix}

We organize the appendix as follows: 

\begin{itemize}
    \item \textbf{Section A}: We provide additional experimental results on VLM continual learning tasks, including evaluations on the more challenging X-TAIL setting, comparisons across different dataset orders, and performance on dataset subsets and their complements.
    
    \item \textbf{Section B}: We validate the generalizability of our approach to LLM-based continual learning tasks using a decoder-only transformer architecture.
    
    \item \textbf{Section C--G}: We conduct detailed analyses of our method, including comparisons with other LoRA-based methods, ablation studies on component transferability, and visualization of activation maps.
    
    \item \textbf{Section H}: We further analyze our method theoretically, focusing on the role of AGO loss, expert pool selection, and the stability of expert merging.
    
    \item \textbf{Section I}: We elaborate on the reproducibility of our experiments and provide all code for public access.
\end{itemize}

\section{More VLM CL Tasks Results}

\subsection{Results on X-TAIL benchmark}

To enable a more comprehensive comparison, we incorporated another powerful benchmark, X-TAIL, from recent research and present the experimental results. In the X-TAIL scenario, we define $\mathcal{C}_{seen}$ as the aggregation of classes from datasets that have been previously learned, and $\mathcal{C}_{unseen}$ as the aggregation of classes that remain unseen. These two sets are disjoint, i.e., $\mathcal{C}_{seen}\cap\mathcal{C}_{unseen}=\varnothing$. The test-time class set encompasses the union of all classes from both seen and unseen datasets, denoted as $\mathcal{C}=\mathcal{C}_{seen}\cup\mathcal{C}_{unseen}$. During evaluation, both in-domain classification and out-of-domain interference must be considered, posing a greater challenge.

The results under the X-TAIL setting are presented in Table \ref{tab:comparison2}. Despite the increased difficulty of this setting leading to reduced performance across all metrics, our method still achieves state-of-the-art results. Specifically, without task-ID information, our approach outperforms the previous SOTA methods by 2.7 percentage points in Transfer, 5.2 points in Average, and 6.3 points in Last. When task-ID is provided, we observe improvements of 0.6, 1.2, and 1.1 percentage points respectively in these metrics. Notably, even under this challenging setup, our method demonstrates exceptional generalization ability that surpasses the zero-shot upper bound performance.
\subsection{Comparison with State-Of-The-Art Methods on Order 2}
The test metrics for Order 2 under the MTIL and X-TAIL settings are shown in Table \ref{tab:mtilo2} and Table \ref{tab:xtailo2}. It can be seen that changes in the dataset order basically have no impact on the training of our method and the stable performance improvement. Our method still achieves state-of-the-art results in all metrics. When prior knowledge of the task-id is available, our method further improves, which is in line with the results in the main paper.

\subsection{Results on Multiple Subsets}
To demonstrate the robustness and generalizability of our method, which is not affected by the number and differences of datasets, we conducted experiments on dataset subsets (randomly selecting 6 out of the 11 datasets: CIFAR100, Flowers, DTD, MNIST, Pets, Aircraft, with the remaining as the complement set: EuroSAT, Caltech101, Cars, SUN397, Food), comparing against multiple baselines: ZSCL, MoE-a, LoRAMoE and more ablations. As can be seen from the Table \ref{tab:subset}, our method is not affected by specific datasets. For both randomly selected subsets and complement sets, our method exhibits the best performance. Additionally, each of the two components achieves positive effects when used independently, and their synergistic combination yields optimal results.
\setlength{\tabcolsep}{4pt}
\begin{table}[h]

\begin{tabular}{lcccccc}
\toprule
\multirow{2}{*}{Method} & \multicolumn{3}{c}{Subset} & \multicolumn{3}{c}{Complement} \\
\cmidrule(lr){2-4} \cmidrule(lr){5-7}
 & Trans. & Avg. & Last & Trans. & Avg. & Last \\
\midrule
\multicolumn{7}{l}{\textbf{Baseline}} \\
ZSCL & 52.7 & 60.6 & 66.8 & 74.5 & 77.4 & 78.3 \\
MoE-a & 57.7 & 64.8 & 70.9 & 76.7 & 78.4 & 79.9 \\
LoRAMoE & 55.4 & 63.2 & 67.2 & 75.0 & 77.1 & 76.9 \\
\midrule
\multicolumn{7}{l}{\textbf{Ablation}} \\
Vanilla LoRA & 54.0 & 61.0 & 65.9 & 72.1 & 74.0 & 74.4 \\
+ Expert Pool & 57.5 & 65.4 & 70.1 & 75.3 & 78.5 & 79.5 \\
+ AGO loss & 54.9 & 62.0 & 67.2 & 73.4 & 75.7 & 76.1 \\
\rowcolor{gray!30} 
+ \textbf{Ours} & \textbf{58.4} & \textbf{67.6} & \textbf{72.3} & \textbf{77.1} & \textbf{80.6} & \textbf{81.7} \\
\bottomrule
\end{tabular}
\caption{Experimental results on dataset subsets and complements under MTIL setting} 
\label{tab:subset}
\end{table}
\subsection{Complete Matrices of Our Method Results}
We present the complete test metrics under the MTIL and X-TAIL settings for Order 1 in  Table \ref{tab:10} and Table \ref{tab:11}. After training on each dataset, tests are conducted on all datasets, each row tracks the model's performance across all datasets after being trained on the corresponding task,  allowing for a thorough assessment of knowledge acquisition and retention. From the results, we can find that our method demonstrates a relatively stable and consistent performance improvement throughout the continual learning sequence. 

\section{Additional Experiments on LLM CL tasks}

\subsection{Experimental Setup}
To further validate the generalizability of our framework on different model architectures, we conducted additional continual fine-tuning experiments on a decoder-only Large Language Model (LLM) \textbf{LLaMA-7B}.

We evaluated the continual learning performance on a \textbf{standard CL benchmark}. This benchmark consists of four text classification datasets: dbpedia, amazon, yahoo, and ag news. We test on three challenging task sequences, as detailed in Table \ref{tab:llama_orders}. For each dataset, following previous classic continual learning settings, we randomly sampled 1000 instances from each category for training and 7600 remaining instances (randomly sampled from all categories) for evaluation, using accuracy as the performance metric, training for 1 epoch. Following the experience from previous work, we only apply LoRA adaptation to the query (q) and value (v) matrices of the transformer blocks.

For a fair comparison, all LoRA-based methods use a rank of 8. For our method, this corresponds to a pool of 8 rank-1 experts. During the forward pass, we dynamically compose an update by selecting 4 experts, and these top-4 experts are also saved and merged after training. As the LLaMA architecture lacks a dedicated \texttt{[CLS]} token, we use the representation of the default first token as the input to our router, which is used simply for sparsification. The learning rate was set to 1e-4, with all other settings consistent with those in the main paper.

\begin{table}[t]
\centering

\begin{tabular}{ccl}
\toprule
\textbf{Order} & \textbf{Model} & \textbf{Task Sequence} \\
\midrule
1 & LLaMA & dbpedia $\rightarrow$ amazon $\rightarrow$ yahoo $\rightarrow$ ag \\
2 & LLaMA & dbpedia $\rightarrow$ amazon $\rightarrow$ ag $\rightarrow$ yahoo \\
3 & LLaMA & yahoo $\rightarrow$ amazon $\rightarrow$ ag $\rightarrow$ dbpedia \\
\bottomrule
\end{tabular}
\caption{Task sequences used for the LLaMA-7B continual learning benchmark.}
\label{tab:llama_orders}
\end{table}

\subsection{Results and Analysis}

\begin{table}[t]
\centering

\begin{tabular}{lcccc}
\toprule
\textbf{Method} & \textbf{Order-1} & \textbf{Order-2} & \textbf{Order-3} & \textbf{avg.} \\
\midrule
Multi-task FT & 80.0 & 80.0 & 80.0 & 80.0 \\
\hdashline
SeqFT & 18.9 & 24.9 & 41.7 & 28.5 \\
SeqLoRA & 44.6 & 32.7 & 53.7 & 43.7 \\
LwF & 54.4 & 53.1 & 49.6 & 52.3 \\
IncLoRA & 66.0 & 64.9 & 68.3 & 66.4 \\
\midrule
\rowcolor{gray!30}
\textbf{Ours} & \textbf{73.1} & \textbf{72.5} & \textbf{75.0} & \textbf{73.5} \\
\bottomrule
\end{tabular}
\caption{Performance comparison (Accuracy \%) on the LLaMA-7B CL benchmark across three different task orders. Our method is compared against several standard CL baselines.}
\label{tab:llama_results_cl}
\end{table}

The results of LLM continual learning experiment are presented in Table \ref{tab:llama_results_cl}. Here, SeqFT means sequentially trains all model parameters on a series of tasks without regularization or replay, SeqLoRA means sequentially trains a fixed set of LoRA parameters, and IncLoRA means incrementally adds and trains new LoRA parameters for each task in a sequential series. As shown, standard sequential fine-tuning methods like SeqFT and SeqLoRA perform poorly, indicating severe catastrophic forgetting. While established CL methods such as LwF and IncLoRA offer improvements, their effectiveness is still limited.

Our method, however, significantly outperforms all baselines across all three task orders, achieving accuracies of 73.1\% (Order-1), 72.5\% (Order-2), and 75.0\% (Order-3). This demonstrates the superior ability of our \textit{Rank-1 Expert Pool} framework to effectively learn new tasks while alleviating forgetting problem on a challenging decoder-only architecture. The results confirm that our dynamic composition mechanism is a robust and highly effective solution for continual learning in various architectures, including LLMs.

\begin{figure}[H]
  \centering
  
  \includegraphics[width=0.8\linewidth]{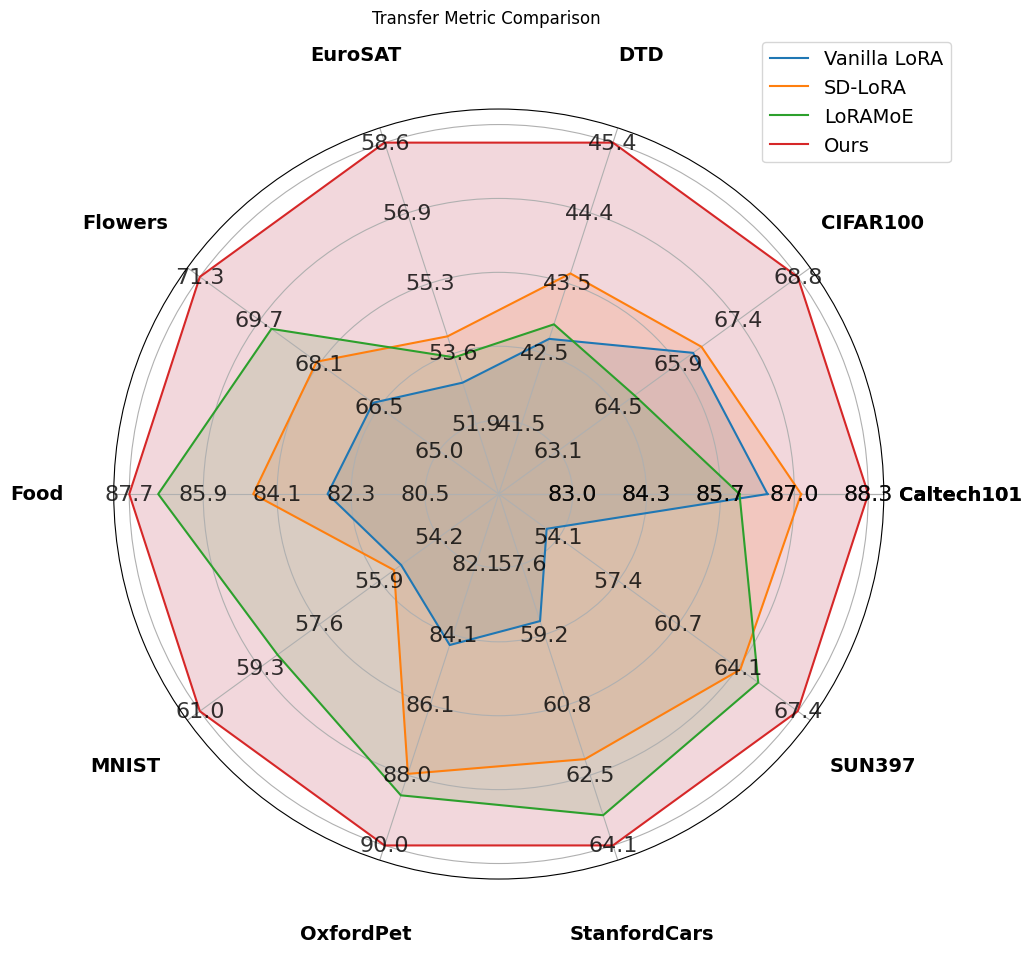}
  \caption{Visualization of the Transfer metrics of more LoRA-based methods on various datasets under the MTIL order I setting.}
  \label{fig:radar}
  
\end{figure}

\section{Detail Comparison with LoRA-Based Method}

To visually highlight the differences in generalization capabilities among various LoRA-based strategies, Figure \ref{fig:radar} presents a comparative analysis of their Transfer metrics across multiple datasets in the MTIL order 1 setting. In this radar chart, a larger and more uniform area covered by a method's plot indicates superior and more stable generalization performance across diverse domains.

As shown, static LoRA variants suffer significant performance drops due to their inability to adapt to dynamic domain shifts in MTIL, resulting in poor generalization. In contrast, our method maintains high performance across datasets, achieving a larger and more balanced area under the curve. This demonstrates its superiority in domain discrimination and parameter isolation, which is the key to effective continual learning. Our dynamic expert composition mechanism enables learning new tasks without interfering with original weights, preserving foundational knowledge.

\section{Plug-and-Play Activation-Guided Orthogonal (AGO) Loss}
To demonstrate the transferability of our proposed \textbf{Activation-Guided Orthogonal (AGO) loss}, we directly apply it as a plug-and-play module to the MoE-a framework, with results in Table \ref{tab:plug}. This is applicable because MoE-a also records activation frequency for its adapters, providing the necessary guidance for our loss. We use this frequency information to apply the orthogonal constraint only to the active adapters for the current task.
The results show that our AGO loss effectively improves the Average and Last metrics of the baseline method (the Transfer metric in MoE-a relies on other components, and is thus unaffected). This experiment highlights a key insight: leveraging activation frequencies to guide a sparse orthogonal constraint is a powerful and generalizable paradigm. It confirms that our AGO loss can be a valuable module for enhancing other dynamic, sparse continual learning frameworks. This indicates that integrating the frequency information of expert selection with the orthogonality in the parameter space is a highly promising direction.
\begin{table}[t]
    \centering
  
    \begin{tabular}{lccc}
        \toprule
        Method & Transfer & Average & Last \\
        \midrule
        MoE-a & 68.9 &71.4 & 76.1 \\
        MoE-a + AGO loss & 68.9 &71.7 & 76.8 \\
        \bottomrule
    \end{tabular}
      \caption{Plug-and-Play ability of Proposed AGO Loss} 
    \label{tab:plug} 
\end{table}
\section{Exploration of VLM Transfer Ability}

\begin{figure}[t]
  \centering
  
  \includegraphics[width=0.91\linewidth]{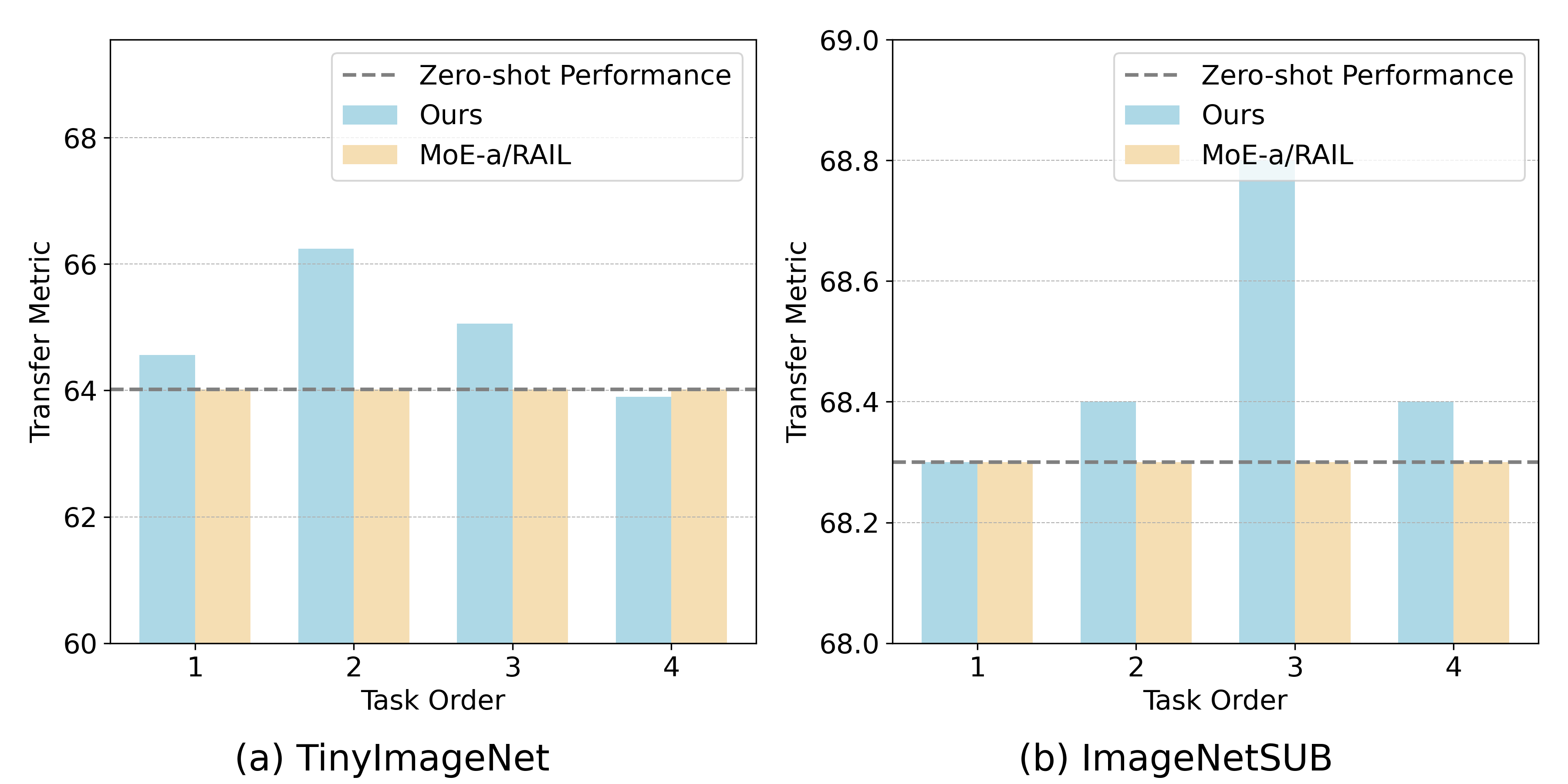}
  \caption{Visualization of Transfer metric of on more datasets. The figure shows that on unseen datasets, our method consistently maintains better generalization ability in continual learning, rather than being limited to the upper bound of CLIP zero-shot ability.}
  \label{fig:more}
  
\end{figure}

As shown in the Figure \ref{fig:more}, during the continual training of the Vision-Language Model, we test it using other different image classification datasets. From the results, it can be observed that our method can effectively preserve the transfer ability of the pre-trained model. Moreover, the classification performance on unseen datasets has even improved, exceeding the zero-shot upper bound performance of CLIP. The results indicate that our method, with the sparsification of selected experts, better maintains the original weight distribution. Additionally, after increasingly aligning more image-text pairs in the training data, it enhances the generalization ability of the pre-trained model more effectively.

\section{Analysis of Rank-1 Expert Pool Configuration and Sparsity}
We analyze the optimal configuration of our Rank-1 Expert Pool, specifically its total size (total rank, $a$), the number of experts selected during the forward pass ($b$), and the number of top experts merged after training ($c$). The results, presented in Table \ref{tab:abc} (formatted as $a$-$b$-$c$), reveal two key findings. First, a sparser final merge (smaller $c$) can enhance the model's generalization ability (Transfer metric). Second, our method demonstrates considerable robustness to these hyperparameters, with performance remaining stable across various configurations. Our final setting is chosen to balance efficiency and overall performance.

\begin{figure*}
  \centering
  
  \includegraphics[width=0.7\linewidth]{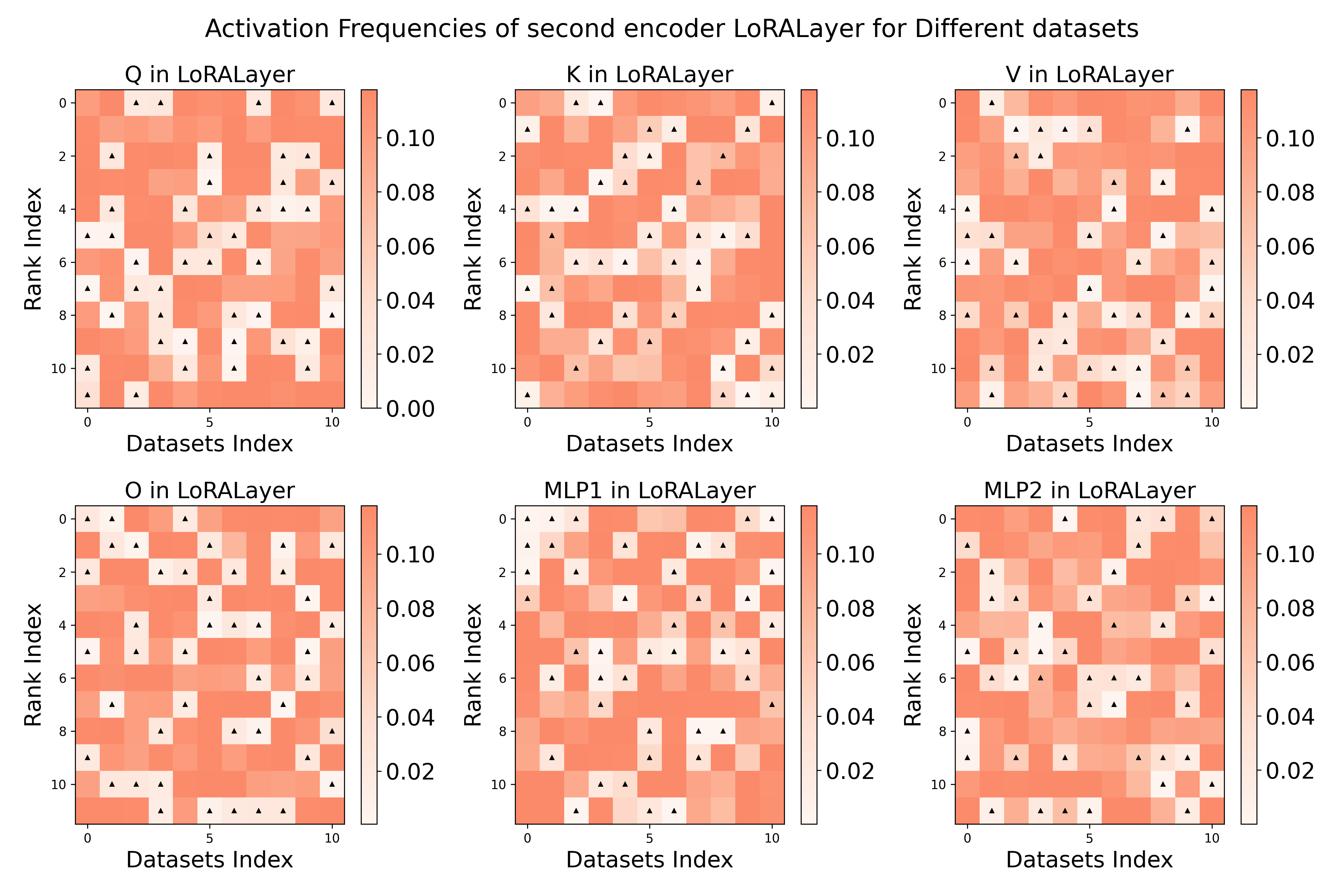}
  \caption{ More Visualization of the activation frequencies of the LoRA layers at the same position of different tasks. The darker the color, the higher the activation frequency. ``$\blacktriangle$'' means disregarded experts. This figure shows frequency information of the second layer in image encoder, indicating that the composed part of each task is exactly the experts with relatively higher activation.}
  \label{fig:act2}
  
\end{figure*}

\begin{figure*}
  \centering
  
  \includegraphics[width=0.7\linewidth]{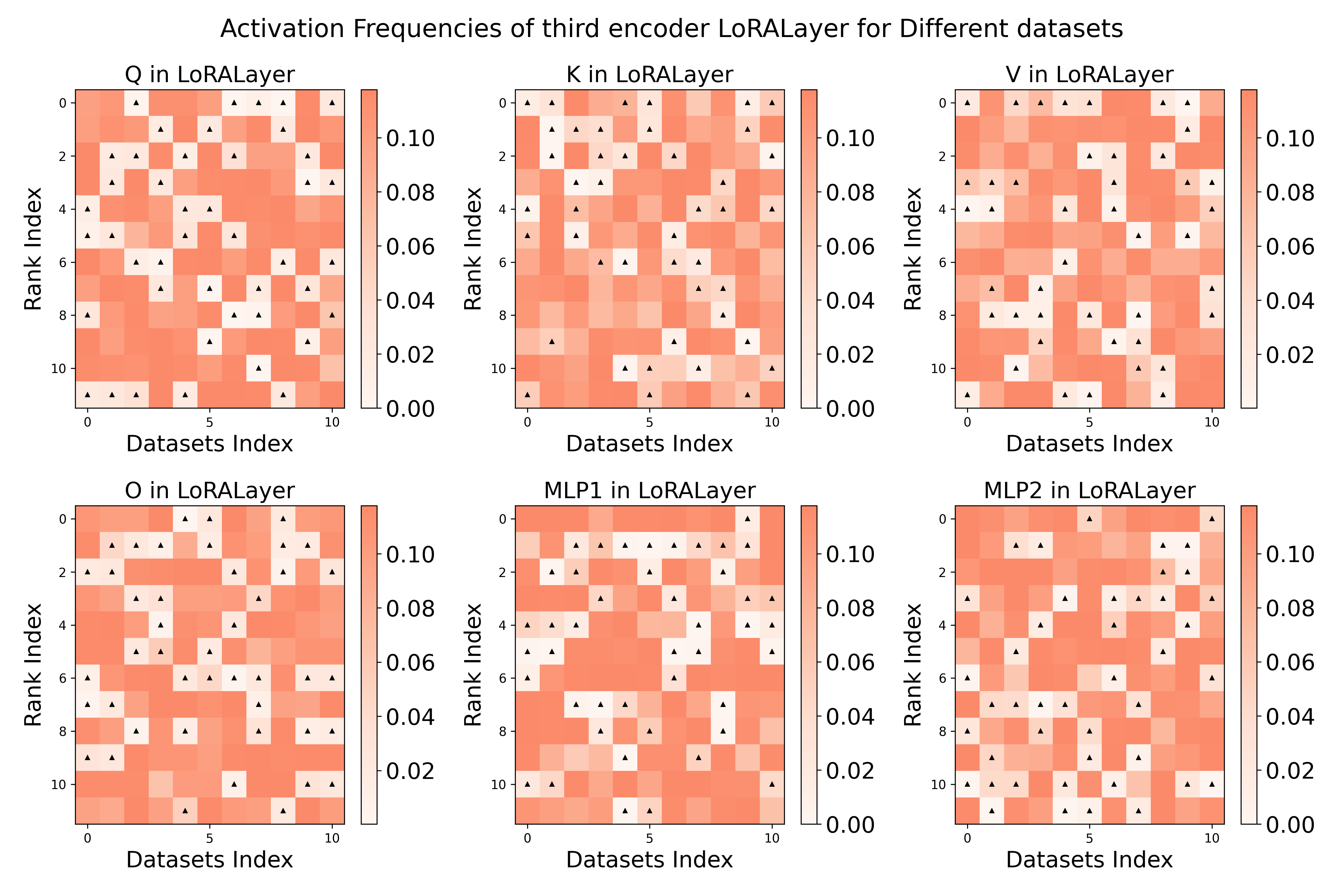}
  \caption{ More Visualization of the activation frequencies of the LoRA layers at the same position of different tasks. The darker the color, the higher the activation frequency. ``$\blacktriangle$'' means disregarded experts. This figure shows frequency information of the third layer in image encoder, indicating that the composed part of each task is exactly the experts with relatively higher activation.}
  \label{fig:act3}
  
\end{figure*}

Moreover, the activation heatmaps in Figures \ref{fig:act2} and \ref{fig:act3} further validate our dynamic composition approach. The visualizations clearly show that our method learns to \textbf{compose} consistent subspaces for data from the same domain (i.e., by selecting similar sets of rank-1 experts), while composing distinctly different subspaces for different domains. This confirms that our framework effectively uses domain characteristics to guide the composition of task-specific updates from the shared expert pool.

\setlength{\tabcolsep}{4pt}
\begin{table}[t]
 
  \begin{tabular}{lcccccc}
    \toprule
    \textbf{Experts Setting} & Trans. & $\Delta$ & Avg. & $\Delta$ & Last & $\Delta$ \\
    \midrule
    \multicolumn{7}{l}{\textbf{Rank-decompose-compose}} \\
    8-4-2 & 70.7 &  \underline{+0.4} & 71.4 &   {-1.6} & 75.3 &   {-2.6} \\
    12-4-2 & 70.0 &   {-0.3} & 72.0 &   {-1.0} & 76.5 &   {-1.4} \\
    \rowcolor{gray!30}
    \textbf{12-8-4 (Ours)} & \textbf{70.3} & 0.0 & \textbf{73.0} & 0.0 & \textbf{77.9} & 0.0 \\
    16-8-4 & 70.3 & 0.0 & 72.2 &   {-0.8} & 76.7 &   {-1.2} \\
    \bottomrule
  \end{tabular}
   \caption{Exploration of the optimal number of experts.}
  \label{tab:abc}
   
\end{table}

\begin{figure*}[h]
  \centering
  \includegraphics[width=0.75\linewidth]{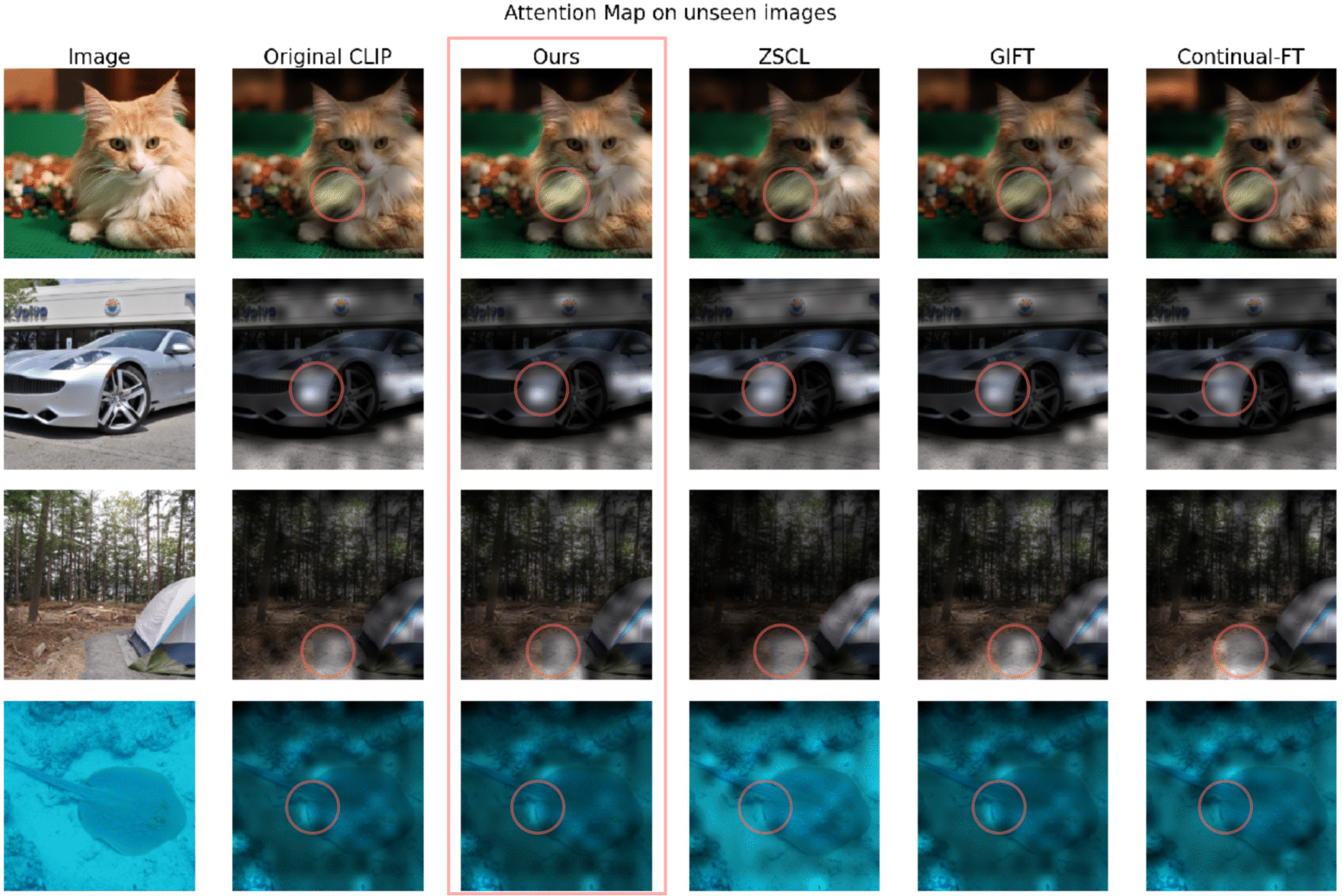}
  \caption{Visualization of the attention maps of different methods on unseen images.
  }
  \label{fig:stack}
  
\end{figure*}

\section{Visualization of Attention Map}
We visualize the attention maps of different CLIP fine-tuning methods on unseen images in Figure \ref{fig:stack}, where the bright parts indicate the areas where attention is concentrated. As illustrated in the figure, in our method, due to the introduction of LoRA weight updates after sparsification, the original weights of CLIP are not significantly affected and basically maintain the original attention. This further ensures the generalization of our method. In contrast, other methods more or less interfere with the original weights, showing different attention maps compared to original CLIP.

\section{Theoretical Analysis}

\subsection{Activation-Guided Orthogonality and Forgetting}

In continual learning, catastrophic forgetting arises from interference between the parameter updates of different tasks. A key strategy to mitigate this is by enforcing orthogonality between their update subspaces. Suppose for task $t$, the LoRA update is represented by its basis vectors $\{b_i\}_{i \in S^{(t)}}$. If these vectors are orthogonal to the basis vectors from all previous tasks (i.e., $b_i^\top b_j \approx 0$), then task-specific knowledge is naturally isolated in the parameter space, preserving what was previously learned.

However, applying this constraint naively to all parameters can be overly restrictive and harm performance on new tasks. We therefore propose the \textbf{Activation-Guided Orthogonal (AGO) loss}, which formalizes this principle in a more targeted and intelligent manner. Our loss operates only on the subspaces defined by the most frequently activated experts. Let $\mathcal{K}_{past}$ be the set of indices for the top-R historically critical experts from all past tasks (derived from their accumulated activation frequencies). Similarly, let $\mathcal{K}_{curr}$ be the set of indices for the top-R experts identified as critical during the training of the current task $t$. The AGO loss is then calculated as:
\[
\mathcal{L}_{AGO} = \frac{1}{|\mathcal{K}_{past}| |\mathcal{K}_{curr}|} \sum_{i \in \mathcal{K}_{past}} \sum_{j \in \mathcal{K}_{curr}} \Big|\left(\mathbf{b}_{past}^i\right)^\top\mathbf{b}_t^j\Big|.
\]
This formulation provides a principled approach to minimizing the probability of parameter collision. By constraining the summation to only the high-frequency expert sets $\mathcal{K}_{past}$ and $\mathcal{K}_{curr}$, the orthogonal pressure is concentrated on the intersection of the most critical subspaces. This encourages the model to find update directions that are \textbf{not only orthogonal but also maximally disjoint} from those of previous tasks, thereby reducing destructive interference on shared basis vectors and improving stability without compromising downstream performance.

\subsection{Sparse Expert Selection and Generalization}

We analyze the benefit of selecting a small set of experts per task. Consider the update rule:
\[
\Delta W^{(t)} = \sum_{i \in S^{(t)}} \pi_i(x) b_i a_i^\top,
\]
where the gating function $\pi_i(x)$ acts as a sparse selector (e.g., a Top-$R$ selector). This induces an implicit $\ell_0$-style regularization on the update space. Formally, given a total pool of $r$ experts, sparse composition reduces the capacity of the updated subspace. This ensures the dimension of the update's basis is much smaller than the total potential rank:
\[
\text{dim}(\text{span}\{b_i\}_{i \in S^{(t)}}) \ll r,
\]
which limits the model's capacity to overfit to the current task and improves generalization on future tasks, as empirically validated by our Transfer metric improvements.

\subsection{Composition Stability and Weight Merging}

We consider the final merged weight update after $T$ tasks:
\[
\hat{W} = W_0 + \alpha \sum_{i \in S^{(1:T)}} \bar{\pi}_i b_i a_i^\top,
\]
where $\bar{\pi}_i$ denotes the averaged activation frequency of expert $i$ and $\alpha$ is a merge scaling factor (set to $1/r$). Under the assumption that the expert subspaces $b_i a_i^\top$ are approximately orthogonal—an assumption encouraged by our \textbf{Activation-Guided Orthogonal (AGO) loss}—the update becomes a stable combination of task-specific directions. The contribution of each expert $j$ to the final weights can be measured by the Frobenius inner product $\langle \cdot, \cdot \rangle$, resulting in:
\[
\langle \hat{W} - W_0, b_j a_j^\top \rangle \propto \bar{\pi}_j.
\]
This formally demonstrates that more critical experts (i.e., those with higher activation frequency $\bar{\pi}_j$) are retained more strongly in the final merged weights. This explains why our strategy of merging the top-$R$ / 2 experts successfully maintains strong performance without needing to store all parameters.

\section{Reproducibility}

To ensure the reproducibility of our research, we have made our code available in supplementary material. All the necessary code can be directly executed to obtain the same results reported in this paper. Upon acceptance of this paper, we commit to fully disclosing all the code and training details, facilitating further exploration and innovation in the field.

\setlength{\tabcolsep}{6pt}

\begin{table*}[t]

    \begin{tabular}{lccccccccccccc}
            \hline    
        \textbf{Method} &\rotatebox{90}{\textbf{Task-ID}} & \rotatebox{90}{\textbf{Aircraft}} & \rotatebox{90}{\textbf{Caltech101}} & \rotatebox{90}{\textbf{CIFAR100}} & \rotatebox{90}{\textbf{DTD}} & \rotatebox{90}{\textbf{EuroSAT}} & \rotatebox{90}{\textbf{Flowers}} & \rotatebox{90}{\textbf{Food}} & \rotatebox{90}{\textbf{MNIST}} & \rotatebox{90}{\textbf{Pets}} & \rotatebox{90}{\textbf{Cars}} & \rotatebox{90}{\textbf{SUN397}} & \rotatebox{90}{\textbf{Average}} \\
        \midrule
        \multicolumn{12}{l}{\textbf{CLIP}} \\
        Zero-shot& \multirow{2}{*}{\rotatebox{90}{\textbf{TIF}}} & 24.3 & 58.6 & 37.2 & 38.6 & 53.8 & 68.9 & 87.9 & 39.6 & 88.7 & 64.6 & 62.4 & 56.8 \\
        5-shot Full Fine-tune & & 29.9 & 62.0 & 59.3 & 53.3 & 88.3 & 87.7 & 85.0 & 84.9 & 80.1 &  67.1 & 68.3 & 69.6 \\
\midrule
        \multicolumn{12}{l}{\textbf{Transfer}} \\
        Continual-FT & \multirow{7}{*}{\rotatebox{90}{\textbf{TIF}}} &-& 52.1 & 25.2 & 19.5 & 15.6 & 23.9 & 44.4 & 20.7 & 34.8 & 17.5 & 41.1 & 29.5  \\
        LwF \small{ {\texttt{[TPAMI'17]}}} & &-& 44.1 & 47.9 & 30.5 & 30.0 & 44.3 & 67.8 & 27.0 & 65.7 & 27.2 & 44.5 & 42.9\\
        LwF-VR \small{ {\texttt{[Arxiv'22]}}} &  &-& 47.1 & 49.5 & 32.8 & 35.0 & 56.4 & 79.6 & 32.5 & 77.5 & 41.9 & 57.7 & 51.0  \\
        WiSE-FT \small{ {\texttt{[CVPR'22]}}} &  &-& 45.0 & 45.1 & 27.5 & 28.1 & 51.1 & 76.6 & 33.3 & 75.3 & 35.8 & 55.8 & 47.4  \\
        ZSCL \small{ {\texttt{[ICCV'23]}}}& &-& 53.3 & 34.6 & 35.4 & 47.9 & 65.3 & \textbf{87.2} & 31.8 & 86.5 & 60.9 & 62.5 & 56.5 \\
        GIFT \small{ {\texttt{[CVPR'25]}}}&  &-& 50.1 & \textbf{55.1} & 35.7 & 41.4 & 61.6 & 84.7 & \textbf{43.6} & 88.2 & 55.5 & \textbf{63.4} &57.9\\
        \rowcolor{gray!30} 
        Ours  &  &-& \textbf{59.0} & 44.0 & \textbf{39.5} & \textbf{54.7} & \textbf{67.5} & 86.9 & 39.3 & \textbf{89.4} & \textbf{63.9} & 61.4 & \textbf{60.6}  \small{(+2.7)}  \\
        \midrule

        MoE-a \small{ {\texttt{[CVPR'24]}}}& \multirow{2}{*}{\rotatebox{90}{ \small{\textbf{TIK}}}} &-& 57.1 & 33.2 & 37.5 & 48.7 & 62.0 & 86.3 & 35.8 & 81.3 & 61.5 & 59.6 & 56.3  \\
        RAIL \small{ {\texttt{[NeurIPS'24]}}}&  &-& 58.6 & 37.2 & 38.6 & 53.8 & \textbf{68.9} & \textbf{87.9} & \textbf{39.6} & 88.7 & \textbf{64.6} & \textbf{62.4} & 60.0  \\
        \rowcolor{gray!30} 
        Ours  &  &-& \textbf{59.0} & \textbf{44.0} & \textbf{39.5} & \textbf{54.7} & 67.5 & 86.9 & 39.3 & \textbf{89.4} & 63.9 & 61.4 & \textbf{60.6}  \small{(+0.6)} \\
        
     \midrule
    \multicolumn{12}{l}{\textbf{Average}} \\
    Continual-FT & \multirow{7}{*}{\rotatebox{90}{\textbf{TIF}}} & 6.1 & \textbf{81.4} & 25.7 & 24.1 & 17.9 & 56.3 & 52.4 & 34.2 & 38.1 & 24.4 & 42.9 & 36.7  \\
    LwF \small{ {\texttt{[TPAMI'17]}}} & & 22.5 & 62.3 & 38.8 & 31.8 & 21.3 & 39.4 & 49.2 & 47.8 & 51.1 & 23.2 & 43.4 & 39.2  \\
    LwF-VR \small{ {\texttt{[Arxiv'22]}}} &  & 26.5 & 55.8 & 57.2 & 43.8 & 41.8 & 70.3 & 79.3 & 43.2 & 77.0 & 44.9 & 58.3 & 54.4   \\
    WiSE-FT \small{ {\texttt{[CVPR'22]}}} &  & 32.0 & 54.4 & 46.0 & 39.3 & 41.1 & 65.5 & 76.8 & 55.6 & 76.4 & 40.9 & 56.8 & 53.2 \\
    ZSCL \small{ {\texttt{[ICCV'23]}}} &  & 30.1 & 51.7 & 48.7 & 43.0 & 64.7 & \textbf{77.8} & \textbf{87.7} & 42.8 & 86.9 & 62.7 & 63.2 &59.9 \\
    GIFT \small{ {\texttt{[CVPR'25]}}} &  & 29.1 & 49.4 & 54.6 & 45.1 & 60.5 & 75.3 & 84.6 & \textbf{59.8} & 87.4 & 57.8 & \textbf{63.8} &60.7\\
    \rowcolor{gray!30} 
    Ours  &  & \textbf{33.4} & 60.6 & \textbf{60.0} & \textbf{54.0} & \textbf{75.8} & 76.2 & 87.3 & 59.5 & \textbf{90.5} & \textbf{65.2} & 61.9 & \textbf{65.9}  \small{(+5.2)} \\
    \midrule
    MoE-a \small{ {\texttt{[CVPR'24]}}} & \multirow{2}{*}{\rotatebox{90}{ \small{\textbf{TIK}}}} & 27.6 & 57.6 & 45.2 & 49.2 & 72.1 & 78.9 & 85.5 & 53.0 & 69.5 & 62.2 & 60.3 & 60.1  \\
    RAIL \small{ {\texttt{[NeurIPS'24]}}} &  & 34.4 & \textbf{81.8} & 51.9 & 50.8 & 60.7 & \textbf{82.6} & \textbf{88.2} & 54.1 & 88.9 & \textbf{66.3} & \textbf{63.1} & 65.7  \\
    \rowcolor{gray!30} 
    Ours  &  & \textbf{36.4} & 63.9 & \textbf{62.2} & \textbf{55.1} & \textbf{76.8} & 76.9 & 87.3 & \textbf{59.5} & \textbf{90.5} & 65.2 & 61.9&\textbf{66.9}  \small{(+1.2)} \\
    
      \midrule
    \multicolumn{12}{l}{\textbf{Last}} \\
    Continual-FT & \multirow{7}{*}{\rotatebox{90}{\textbf{TIF}}} & 8.5 & \textbf{82.4} & 24.6 & 28.9 & 33.6 & 83.9 & 65.5 & 65.0 & 50.7 & 54.1 & 61.3& 50.8  \\
    LwF \small{ {\texttt{[TPAMI'17]}}} &  & 10.5 & 72.2 & 15.8 & 20.2 & 0.1 & 13.7 & 22.0 & 79.2 & 12.8 & 1.7 & 32.3 & 25.5  \\
    LwF-VR \small{ {\texttt{[Arxiv'22]}}} &  & 16.9 & 72.1 & 54.6 & 45.8 & 45.9 & 77.4 & 78.4 & 61.5 & 75.4 & 57.4 & 64.3 &59.1  \\
    WiSE-FT \small{ {\texttt{[CVPR'22]}}} &  & 29.5 & 56.7 & 48.8 & 45.8 & 39.6 & 79.7 & 79.7 & 94.0 & 82.5 & 63.6 & 67.5 & 62.5 \\
    ZSCL \small{ {\texttt{[ICCV'23]}}} &  & 26.9 & 51.3 & 56.7 & 45.6 & 71.5 & \textbf{87.2} & \textbf{88.6} & 64.5 & 89.3 & \textbf{70.9} & \textbf{70.2} & 65.7 \\
    GIFT \small{ {\texttt{[CVPR'25]}}} &  & 27.8 & 49.4 & 32.3 & 47.1 & 58.9 & 83.8 & 83.2 & 87.2 & 83.5 & 67.6 & 67.8&62.6\\
    \rowcolor{gray!30} 
    Ours &  & \textbf{31.1} & 59.6 & \textbf{62.8} & \textbf{57.1} & \textbf{86.2} & 81.6 & 87.9 & \textbf{95.0} & \textbf{93.2} & \textbf{70.9} & 67.2 & \textbf{72.0}  \small{(+6.3)} \\
    \midrule
    MoE-a \small{ {\texttt{[CVPR'24]}}} & \multirow{2}{*}{\rotatebox{90}{ \small{\textbf{TIK}}}} & 27.6 & 53.9 & 48.6 & 53.9 & 86.0 & 92.8 & 84.5 & 80.7 & 41.7 & 65.4 & 67.4 & 63.9  \\
    RAIL \small{ {\texttt{[NeurIPS'24]}}} &  & 34.7 & \textbf{88.2} & 65.2 & 58.4 & 64.7 & \textbf{93.7} & \textbf{88.7} & 78.3 & 89.8 & \textbf{74.2} & \textbf{69.7} & 73.2  \\
    \rowcolor{gray!30} 
    Ours  &  &  \textbf{36.4} & 64.4 & \textbf{66.2} & \textbf{61.0} & \textbf{89.5} & 84.7 & 87.8 & \textbf{95.0} & \textbf{93.5} & 71.2 & 67.2 &\textbf{74.3}  \small{(+1.1)} \\
    
      \midrule
    \end{tabular}
        \caption{Comparisons with state-of-the-art methods on few-shot X-TAIL Order 1 benchmark in terms of ``Transfer'', ``Average'', and ``Last'' scores (\%). \textbf{TIK} means Task-Id Known, and \textbf{TIF} means Task-Id Free. We label the best method with bold style.}
    
    \label{tab:comparison2}
\end{table*}

\begin{table*}

    \renewcommand{\arraystretch}{1} 
    
    \begin{tabular}{lccccccccccccc}
            \hline   
        \textbf{Method} &\rotatebox{90}{\textbf{Task-ID}} & \rotatebox{90}{\textbf{Cars}} & \rotatebox{90}{\textbf{Food}} & \rotatebox{90}{\textbf{MNIST}} & \rotatebox{90}{\textbf{Pets}} & \rotatebox{90}{\textbf{Flowers}} & \rotatebox{90}{\textbf{SUN397}} & \rotatebox{90}{\textbf{Aircraft}} & \rotatebox{90}{\textbf{Caltech101}} & \rotatebox{90}{\textbf{DTD}} & \rotatebox{90}{\textbf{EuroSAT}} & \rotatebox{90}{\textbf{CIFAR100}} & \rotatebox{90}{\textbf{Average}} \\
        \midrule
        \multicolumn{12}{l}{\textbf{CLIP}} \\
       Zero-shot& \multirow{2}{*}{\rotatebox{90}{\textbf{TIF}}} & 64.7 & 88.5 & 59.4 & 89.0 & 71.0 & 65.2 & 24.3 & 88.4 & 44.6 & 54.9 & 68.2 & 65.3 \\
5-shot Full Fine-tune & & 65.4 & 83.3 & 96.6 & 84.9 & 92.9 & 71.3 & 30.6 & 93.5 & 65.1 & 91.7 & 76.8 & 77.5 \\
         \midrule 
        \multicolumn{12}{l}{\textbf{Transfer}} \\
        Continual-FT & \multirow{7}{*}{\rotatebox{90}{\textbf{TIF}}} &-& 76.0 & 64.6 & 67.1 & 49.7 & 53.7 & 8.3 & 77.9 & 33.9 & 23.9 & 37.1 & 49.2  \\
LwF \small{ {\texttt{[TPAMI'17]}}} & &-& 64.2 & 59.1 & 68.1 & 38.4 & 54.9 & 6.7 & 78.0 & 35.5 & 33.5 & 47.4 & 48.6\\
LwF-VR \small{ {\texttt{[Arxiv'22]}}} &  &-& 80.1 & 55.4 & 77.7 & 50.4 & 61.4 & 9.1 & 83.5 & 40.1 & 31.5 & 54.8 & 54.4  \\
WiSE-FT \small{ {\texttt{[CVPR'22]}}} &  &-& 77.3 & 60.0 & 76.9 & 54.2 & 58.0 & 11.1 & 81.8 & 37.6 & 31.7 & 48.1 & 53.7  \\
ZSCL \small{ {\texttt{[ICCV'23]}}}& &-& 87.3 & \textbf{64.8} & 85.3 & 67.9 & 64.5 & 18.9 & 86.6 & 43.6 & 43.2 & 65.7 & 62.8 \\
        GIFT \small{ {\texttt{[CVPR'25]}}}&  & -& 87.8 & 62.0 & 88.4 & 66.7 & \textbf{68.6} & 20.1 & \textbf{88.9} & 46.0 & 50.5 & 65.7 & 64.5 \\
        \rowcolor{gray!30} 
        
        Ours  &  &-& \textbf{88.8} & 57.6 & \textbf{89.8} & \textbf{69.1} & 66.0 & \textbf{24.0} & 88.8 & \textbf{46.6} & \textbf{53.5} & \textbf{68.3} & \textbf{65.3}  \small{(+0.8)} \\
        \midrule

        MoE-a \small{ {\texttt{[CVPR'24]}}}& \multirow{2}{*}{\rotatebox{90}{ \small{\textbf{TIK}}}} &-& \textbf{88.8} & \textbf{59.5} & 89.1 & \textbf{71.2} & 65.3 & 18.2 & 87.9 & 44.2 & 54.6 & 68.2 & 64.7  \\
        RAIL \small{ {\texttt{[NeurIPS'24]}}}&  &- & 88.5 & 59.4 & 89.0 & 71.0 & 65.2 & \textbf{24.3} & 88.4 & 44.6 & \textbf{54.9} & 68.2 & \textbf{65.3} \\
\rowcolor{gray!30} 
        
        Ours &  &-& \textbf{88.8} & 57.6 & \textbf{89.8} & 69.1 & \textbf{66.0} & 24.0 & \textbf{88.8} & \textbf{46.6} & 53.5 & \textbf{68.3} & \textbf{65.3} \small{(+0.0)}\\
        
       \midrule
    \multicolumn{12}{l}{\textbf{Average}} \\
    Continual-FT & \multirow{7}{*}{\rotatebox{90}{\textbf{TIF}}} & 50.1 & 56.9 & 73.5 & 64.5 & 45.9 & 51.2 & 8.2 & 81.8 & 37.9 & 29.9 & 38.6 & 49.0  \\
LwF \small{ {\texttt{[TPAMI'17]}}} & & 64.1 & 55.0 & 79.5 & 69.2 & 55.7 & 58.3 & 10.8 & 81.7 & 41.3 & 39.2 & 47.4 & 54.7  \\
LwF-VR \small{ {\texttt{[Arxiv'22]}}} &  & 63.3 & 76.9 & 71.4 & 79.1 & 68.9 & 65.0 & 13.4 & 86.0 & 45.7 & 36.3 & 55.3 & 60.1   \\
WiSE-FT \small{ {\texttt{[CVPR'22]}}} &  & 59.3 & 64.7 & 77.4 & 70.3 & 51.3 & 58.6 & 10.8 & 84.2 & 42.0 & 38.6 & 49.1 & 55.1 \\
ZSCL \small{ {\texttt{[ICCV'23]}}} &  & 70.0 & 85.0 & 79.8 & 86.1 & 79.4 & 68.3 & 21.8 & 88.8 & 48.8 & 49.3 & 66.5 & 67.6 \\
        GIFT \small{ {\texttt{[CVPR'25]}}} &  & 68.0 & 85.1 & 85.7 & 88.6 & 79.9 & \textbf{71.2} & 24.7 & \textbf{89.9} & 50.5 & 55.5 & 66.4 & 69.6 \\
        \rowcolor{gray!30} 
        
        Ours  &  &\textbf{71.4} & \textbf{88.8} & \textbf{88.0} & \textbf{92.5} & \textbf{80.8} & 69.5 & \textbf{29.7} & 89.6 & \textbf{52.0} & \textbf{60.0} & \textbf{69.0} & \textbf{71.9}  \small{(+2.3)}\\
        \midrule
        
    MoE-a \small{ {\texttt{[CVPR'24]}}} & \multirow{2}{*}{\rotatebox{90}{ \small{\textbf{TIK}}}} & 61.2 & 87.0 & 87.3 & 89.1 & 79.3 & 68.5 & 23.4 & 89.4 & 49.9 & 60.8 & 68.8 & 69.5  \\
        RAIL \small{ {\texttt{[NeurIPS'24]}}} &  & 66.1 & 88.5 & 70.4 & 89.0 & \textbf{84.4} & 65.7 & \textbf{32.9} & \textbf{94.5} & \textbf{58.1} & \textbf{71.8} & 68.2 & 71.9  \\

        \rowcolor{gray!30} 
        
        Ours &  & \textbf{73.9} & \textbf{89.8} & \textbf{88.2} & \textbf{92.9} & 82.8 & \textbf{69.7} & 30.2 & 89.7 & 52.2 & 60.2 & \textbf{69.0}& \textbf{72.6}  \small{(+0.7)} \\
        
   \midrule
    \multicolumn{12}{l}{\textbf{Last}} \\
    Continual-FT & \multirow{7}{*}{\rotatebox{90}{\textbf{TIF}}} & 35.2 & 28.6 & 58.3 & 51.2 & 14.0 & 46.1 & 5.3 & 89.5 & 47.0 & 52.9 & 53.6 & 42.8  \\
LwF \small{ {\texttt{[TPAMI'17]}}} &  & 57.1 & 40.1 & 84.1 & 58.1 & 50.5 & 57.6 & 14.3 & 87.9 & 54.7 & 64.0 & 47.0 & 56.8  \\
LwF-VR \small{ {\texttt{[Arxiv'22]}}} &  & 57.3 & 70.1 & 72.1 & 74.6 & 71.9 & 65.8 & 17.4 & 89.5 & 60.0 & 56.0 & 60.2 & 63.5  \\
WiSE-FT \small{ {\texttt{[CVPR'22]}}} &  & 48.1 & 47.7 & 66.9 & 59.8 & 25.0 & 56.1 & 7.4 & 88.5 & 52.2 & 66.8 & 59.4 & 51.8 \\
ZSCL \small{ {\texttt{[ICCV'23]}}} &  & 67.4 & 82.7 & 78.7 & 85.7 & 81.3 & 71.2 & 25.0 & \textbf{92.5} & 62.0 & 72.2 & 74.4 & 71.8 \\
        GIFT \small{ {\texttt{[CVPR'25]}}} &  & 67.6 & 82.8 & 88.8 & 89.0 & 83.6 & \textbf{73.3} & 30.8 & 91.5 & 62.7 & 71.9 & 74.3 & 74.2 \\
        \rowcolor{gray!30} 
        
        Ours &  & \textbf{69.3} & \textbf{87.7} & \textbf{94.2} & \textbf{93.0} & \textbf{84.3} & 72.5 & \textbf{35.1} & 91.2 & \textbf{66.0} & \textbf{87.8} & \textbf{76.5}& \textbf{78.0}  \small{(+3.8)} \\
        \midrule
        
    MoE-a \small{ {\texttt{[CVPR'24]}}} & \multirow{2}{*}{\rotatebox{90}{ \small{\textbf{TIK}}}} & 59.4 & 87.0 & 91.8 & 89.0 & 84.1 & 71.9 & 29.4 & 91.4 & 64.2 & 88.8 & 75.0 & 75.7  \\
        RAIL \small{ {\texttt{[NeurIPS'24]}}} &  & 72.5 & 88.5 & 89.7 & 89.1 & 89.0 & \textbf{95.6} & 32.9 & \textbf{95.1} & 63.2 & 81.5 & 70.3 & 77.2  \\

        \rowcolor{gray!30} 
        
        Ours &  &\textbf{73.8} & \textbf{89.9} & \textbf{95.0} & \textbf{94.1} & \textbf{90.6} & 72.8 & \textbf{37.6} & 91.1 & \textbf{67.1} & \textbf{90.4} & \textbf{76.5} & \textbf{79.9}  \small{(+2.7)} \\
      \midrule
    \end{tabular}
        \caption{Comparisons with state-of-the-art methods on few-shot MTIL Order 2 benchmark in terms of ``Transfer'', ``Average'', and ``Last'' scores (\%). \textbf{TIK} means Task-Id Known, and \textbf{TIF} means Task-Id Free. We label the best method with bold style.}
    \label{tab:mtilo2}
\end{table*}

\begin{table*}
   
    \renewcommand{\arraystretch}{1} 
    \begin{tabular}{lccccccccccccc}
            \hline     
        \textbf{Method} &\rotatebox{90}{\textbf{Task-ID}} & \rotatebox{90}{\textbf{Cars}} & \rotatebox{90}{\textbf{Food}} & \rotatebox{90}{\textbf{MNIST}} & \rotatebox{90}{\textbf{Pets}} & \rotatebox{90}{\textbf{Flowers}} & \rotatebox{90}{\textbf{SUN397}} & \rotatebox{90}{\textbf{Aircraft}} & \rotatebox{90}{\textbf{Caltech101}} & \rotatebox{90}{\textbf{DTD}} & \rotatebox{90}{\textbf{EuroSAT}} & \rotatebox{90}{\textbf{CIFAR100}} & \rotatebox{90}{\textbf{Average}} \\
            \hline    
        \multicolumn{12}{l}{\textbf{CLIP}} \\
        Zero-shot& \multirow{2}{*}{\rotatebox{90}{\textbf{TIF}}} & 64.6 & 87.9 & 39.6 & 88.7 & 68.9 & 62.4 & 24.3 & 58.6 & 38.6 & 53.8 & 37.2 & 56.8 \\
        5-shot Full Fine-tune & & 67.1 & 85.0 & 84.9 & 80.1 & 87.7 & 68.3 & 29.9 & 62.0 & 53.3 & 88.3 & 59.3 & 69.6 \\
        \midrule
        \multicolumn{12}{l}{\textbf{Transfer}} \\
        Continual-FT & \multirow{7}{*}{\rotatebox{90}{\textbf{TIF}}} &-& 87.0 & 40.3 & 79.5 & 55.5 & 57.1 & 15.2 & 47.4 & 26.6 & 20.4 & 25.7 & 45.5  \\
        LwF \small{ {\texttt{[TPAMI'17]}}} & &-& 85.9 & 40.2 & 69.7 & 43.7 & 49.6 & 9.4 & 54.8 & 26.5 & 18.3 & 25.4 & 42.4\\
        LwF-VR \small{ {\texttt{[Arxiv'22]}}} &  &-& 87.4 & 42.7 & 79.2 & 57.6 & 59.5 & 14.2 & 49.8 & 31.8 & 33.7 & 47.8 & 50.4  \\
        WiSE-FT \small{ {\texttt{[CVPR'22]}}} &  &-& 86.5 & 36.6 & 80.3 & 59.2 & 59.6 & 16.8 & 48.7 & 28.5 & 22.4 & 35.7 & 47.4  \\
        ZSCL \small{ {\texttt{[ICCV'23]}}} &  &-& 87.5 & 33.6 & 85.5 & \textbf{66.0} & 62.4 & 20.3 & 51.4 & 36.3 & 47.4 & 43.6 & 53.4 \\
        GIFT \small{ {\texttt{[CVPR'25]}}} &  &-& 87.4 & \textbf{46.0} & 88.1 & 63.8 & \textbf{65.6} & 20.1 & 50.1 & 36.7 & 41.2 & \textbf{55.4} &55.4\\
        \rowcolor{gray!30} 
        Ours  &  &- & \textbf{87.6} & 38.7 & \textbf{89.6} & 65.4 & 62.3 & \textbf{23.7} & \textbf{58.7} & \textbf{41.1} & \textbf{48.4} & 51.8 & \textbf{56.7}  \small{(+1.3)}  \\
                \midrule

        MoE-a \small{ {\texttt{[CVPR'24]}}} & \multirow{2}{*}{\rotatebox{90}{ \small{\textbf{TIK}}}} &-& 87.2 & \textbf{41.2} & 87.9 & 64.5 & 61.6 & 18.7 & 53.3 & 35.4 & 43.6 & 36.0 & 52.9  \\
        
        RAIL \small{ {\texttt{[NeurIPS'24]}}} &  &-& \textbf{87.9} & 39.6 & 88.7 & \textbf{68.9} & \textbf{62.4} & \textbf{24.3} & 58.6 & 38.6 & \textbf{53.8} & 37.2 & 56.0  \\

        \rowcolor{gray!30} 
        Ours  &  &-& 87.6 & 38.7 & \textbf{89.6} & 65.4 & 62.3 & 23.7 & \textbf{58.7} & \textbf{41.1} & 48.4 & \textbf{51.8} & \textbf{56.7}  \small{(+0.7)} \\
    \midrule
    \multicolumn{12}{l}{\textbf{Average}} \\
    Continual-FT & \multirow{7}{*}{\rotatebox{90}{\textbf{TIF}}} & 41.6 & 61.3 & 53.3 & 61.0 & 46.7 & 53.5 & 11.4 & 62.1 & 26.9 & 22.6 & 26.0 & 42.4  \\
    LwF \small{ {\texttt{[TPAMI'17]}}} &  & 30.5 & 45.6 & 77.9 & 40.7 & 40.1 & 48.0 & 7.6 &\textbf{64.7}& 26.8 & 16.8 & 25.2 & 38.5  \\
    LwF-VR \small{ {\texttt{[Arxiv'22]}}} &  & 56.2 & 80.6 & 47.7 & 71.4 & 72.2 & 63.1 & 16.9 & 55.3 & 36.5 & 33.6 & 48.7& 52.9   \\
    WiSE-FT \small{ {\texttt{[CVPR'22]}}} &  & 45.6 & 70.5 & 69.4 & 62.4 & 52.4 & 56.7 & 13.3 & 62.9 & 30.5 & 25.7 & 36.1 & 47.8 \\
    ZSCL \small{ {\texttt{[ICCV'23]}}} &  & 69.5 & \textbf{88.1} & 60.9 & 87.7 & \textbf{80.2} & 66.7 & 23.8 & 51.9 & 39.7 & 53.0 & 45.3 &60.6 \\
    GIFT \small{ {\texttt{[CVPR'25]}}} &  & 68.0 & 84.9 & 80.6 & 87.0 & 77.3 & \textbf{68.1} & 24.6 & 51.4 & 40.3 & 45.5 & \textbf{56.3} &62.2\\
    \rowcolor{gray!30} 
        
    Ours  &  & \textbf{72.0} & 87.8 & \textbf{84.1} & \textbf{92.3} & 75.8 & 65.3 & \textbf{29.0} & 60.7 & \textbf{46.5} & \textbf{55.7} & 52.8 & \textbf{65.6}  \small{(+3.4)} \\
            \midrule
    
    MoE-a \small{ {\texttt{[CVPR'24]}}} & \multirow{2}{*}{\rotatebox{90}{ \small{\textbf{TIK}}}} & 60.1 & 83.1 & 51.0 & 78.9 & 77.4 & 63.3 & 19.2 & 51.4 & 38.6 & 43.6 & 37.1 & 54.9  \\
    RAIL \small{ {\texttt{[NeurIPS'24]}}} &  & 72.9 & 88.4 & 47.7 & 89.4 & \textbf{83.5} & 65.7 & \textbf{29.1} & \textbf{65.6} & 44.0 & \textbf{57.4} & 39.7 & 62.1  \\

    \rowcolor{gray!30} 
        
    Ours  &  & \textbf{74.1} &\textbf{ 89.1} & \textbf{84.5} & \textbf{92.8} & 78.2 & \textbf{66.1} & \textbf{29.1} & 60.7 & \textbf{46.8} & 55.8 & \textbf{52.8}&\textbf{66.4}  \small{(+4.3)} \\
        
 \midrule
    \multicolumn{12}{l}{\textbf{Last}} \\
    Continual-FT & \multirow{7}{*}{\rotatebox{90}{\textbf{TIF}}} & 13.6 & 32.5 & 11.5 & 29.5 & 8.5 & 45.9 & 1.7 & \textbf{88.9} & 25.0 & 27.3 & 28.7& 28.5  \\
    LwF \small{ {\texttt{[TPAMI'17]}}} &  & 5.7 & 32.6 & 82.3 & 10.8 & 18.1 & 35.8 & 2.6 & 83.9 & 30.5 & 10.0 & 23.3 & 30.5  \\
    LwF-VR \small{ {\texttt{[Arxiv'22]}}} &  & 42.0 & 73.6 & 44.0 & 48.8 & 67.2 & 62.5 & 16.3 & 69.4 & 48.5 & 30.7 & 57.2 &50.9  \\
    WiSE-FT \small{ {\texttt{[CVPR'22]}}} &  & 16.0 & 61.2 & 59.8 & 33.0 & 25.9 & 50.5 & 5.9 & 88.5 & 33.0 & 38.3 & 39.9 & 41.1 \\
    ZSCL \small{ {\texttt{[ICCV'23]}}} &  &67.0 & \textbf{87.6} & 65.0 & 87.5 & \textbf{87.0} & 70.0 & 26.8 & 52.6 & 48.1 & 76.7 & 62.3 & 66.4 \\
    GIFT \small{ {\texttt{[CVPR'25]}}} &  & 67.5 & 82.7 & 85.0 & 87.9 & 81.0 & \textbf{70.3} & 30.6 & 50.7 & 48.1 & 57.2 & \textbf{65.1}&66.0\\
    \rowcolor{gray!30} 
    Ours &  & \textbf{69.7} & 86.5 & \textbf{93.3} & \textbf{92.3} & 79.2 & 67.0 & \textbf{35.3} & 63.0 & \textbf{59.5} & \textbf{88.0} & 62.2 & \textbf{72.4}  \small{(+6.0)} \\
    \midrule
    
    MoE-a \small{ {\texttt{[CVPR'24]}}} & \multirow{2}{*}{\rotatebox{90}{ \small{\textbf{TIK}}}} & 60.8 & 84.2 & 65.9 & 75.3 & 84.9 & 64.1 & 19.7 & 47.6 & 50.7 & 31.1 & 47.4 & 57.4  \\
    RAIL \small{ {\texttt{[NeurIPS'24]}}} &  & 73.2 & 88.7 & 66.2 & 89.8 & \textbf{93.7} &\textbf{ 69.6} & 35.0 & \textbf{89.1} & 59.6 & 74.2 & \textbf{65.1} & 73.1  \\

    \rowcolor{gray!30} 
        
    Ours  &  & \textbf{74.1} & \textbf{89.3} & \textbf{94.7} & \textbf{94.0} & 85.4 & 69.3 & \textbf{35.6} & 64.1 & \textbf{62.0} & \textbf{89.2} & 62.2 &\textbf{74.5}  \small{(+1.4)} \\
  \midrule
    \end{tabular}
     \caption{Comparisons with state-of-the-art methods on few-shot X-TAIL Order 2 benchmark in terms of ``Transfer'', ``Average'', and ``Last'' scores (\%). \textbf{TIK} means Task-Id Known, and \textbf{TIF} means Task-Id Free. We label the best method with bold style.}
    \label{tab:xtailo2}
\end{table*}

\setlength{\tabcolsep}{3pt}

\begin{table*}
  
    \renewcommand{\arraystretch}{0.9} 
\begin{tabular}{lcccccccccccc}
        \toprule
        & Aircraft & Caltech101 & CIFAR100 & DTD & EuroSAT & Flowers & Food & MNIST & OxfordPet & Cars & SUN397 \\
        \midrule
        \multicolumn{12}{l}{\textbf{Transfer}} \\
        Transfer &  & 88.3 & 68.8 & 45.4 & 58.6 & 71.3 & 87.7 & 61.0 & 90.0 & 64.1 & 67.4 & \textbf{70.3} \\
        \midrule
        \multicolumn{12}{l}{\textbf{Task Performance}} \\
        Aircraft & \cellcolor{gray!30}36.51 & 88.31 & 68.22 & 45.11 & 54.87 & 72.09 & 88.97 & 58.86 & 90.05 & 65.25 & 64.82 &-\\
        Caltech101 & 36.6 & \cellcolor{gray!30}90.9 & 69.46 & 45.74 & 58.93 & 72.42 & 88.94 & 58.78 & 89.78 & 65.53 & 67.43 &-\\
        CIFAR100 & 34.11 & 91.47 & \cellcolor{gray!30}77.9 & 45.43 & 61.87 & 71.3 & 88.41 & 60.26 & 90.02 & 64.76 & 68.76 &-\\
        DTD & 34.26 & 90.84 & 77.31 & \cellcolor{gray!30}65.27 & 58.63 & 70.94 & 86.86 & 62.07 & 90.43 & 64.23 & 68.01 &-\\
        EuroSAT & 33.18 & 90.5 & 76.39 & 65.48 & \cellcolor{gray!30}90.11 & 69.69 & 86.43 & 62.51 & 89.48 & 63.65 & 67.57 &-\\
        Flowers & 33.48 & 91.13 & 76.26 & 65.8 & 89.69 & \cellcolor{gray!30}88.45 & 86.84 & 61.65 & 90.62 & 64.12 & 67.62 &-\\
        Food & 33.63 & 92.17 & 76.48 & 65.37 & 89.44 & 87.97 & \cellcolor{gray!30}88.39 & 63.04 & 90.11 & 63.91 & 67.78 &-\\
        MNIST & 32.37 & 92.17 & 74.25 & 64.57 & 89.24 & 87.56 & 88.18 & \cellcolor{gray!20}95.17 & 89.15 & 62.93 & 67.6 &-\\
        OxfordPet & 32.01 & 92.51 & 74.45 & 64.57 & 88.0 & 87.38 & 88.4 & 95.13 & \cellcolor{gray!30}93.73 & 62.65 & 67.39 &-\\
        Cars & 30.45 & 92.45 & 74.33 & 64.63 & 87.76 & 86.88 & 88.36 & 94.96 & 89.5 & \cellcolor{gray!30}71.45 & 67.3 &-\\
        SUN397 & 31.11 & 93.03 & 74.96 & 64.1 & 87.37 & 85.41 & 88.49 & 95.13 & 93.46 & 71.22 & \cellcolor{gray!30}72.81 & \textbf{77.9} \\
        \hdashline
        \multicolumn{12}{l}{\textbf{Average}} \\
        Average & 33.4 & 91.4 & 74.5 & 59.6 & 77.8 & 80.0 & 88.0 & 73.4 & 90.9 & 65.4 & 67.9 & \textbf{73.0} \\
        \bottomrule
    \end{tabular}
      \caption{Accuracy (\%) of our method (Ours) on the MTIL benchmark with order-I. Each row represents the performance on every dataset of the model trained after the corresponding task. \textit{Transfer}, \textit{Average}, and \textit{Last} metrics are shown with bold style.}
    \label{tab:10}
\end{table*}

\begin{table*}
  
    \renewcommand{\arraystretch}{0.9} 
\begin{tabular}{lcccccccccccc}
        \toprule
        & Aircraft & Caltech101 & CIFAR100 & DTD & EuroSAT & Flowers & Food & MNIST & OxfordPet & Cars & SUN397 \\
        \midrule
        \multicolumn{12}{l}{\textbf{Transfer}} \\
        Transfer &  & 59.0 & 44.0 & 39.5 & 54.7 & 67.5 & 86.9 & 39.3 & 89.4 & 63.9 & 61.4 & \textbf{60.6} \\
        \midrule
        \multicolumn{12}{l}{\textbf{Task Performance}} \\
        Aircraft & \cellcolor{gray!30}36.45 & 58.99 & 39.86 & 39.2 & 53.33 & 69.51 & 88.33 & 36.0 & 89.7 & 65.22 & 61.67 &-\\
        Caltech101 & 36.57 & \cellcolor{gray!30}64.4 & 48.08 & 39.95 & 56.26 & 69.23 & 88.42 & 35.67 & 89.53 & 65.46 & 62.74 &-\\
        CIFAR100 & 34.08 & 60.37 & \cellcolor{gray!30}66.25 & 39.26 & 57.98 & 67.07 & 87.84 & 39.58 & 89.53 & 64.63 & 62.99 &-\\
        DTD & 34.23 & 60.2 & 66.09 & \cellcolor{gray!30}61.01 & 51.13 & 66.48 & 85.76 & 40.21 & 89.81 & 63.91 & 61.28 &-\\
        EuroSAT & 33.15 & 59.27 & 64.37 & 60.74 & \cellcolor{gray!30}89.52 & 65.15 & 85.27 & 41.0 & 88.83 & 63.39 & 60.82 &-\\
        Flowers & 33.45 & 60.48 & 64.64 & 60.85 & 88.91 & \cellcolor{gray!30}84.7 & 85.98 & 40.21 & 89.89 & 63.84 & 60.93 &-\\
        Food & 33.6 & 60.89 & 65.02 & 60.11 & 88.61 & 84.19 & \cellcolor{gray!30}87.75 & 42.41 & 89.53 & 63.65 & 61.23 &-\\
        MNIST & 32.34 & 60.77 & 60.86 & 58.46 & 88.43 & 83.74 & 87.67 & \cellcolor{gray!30}94.99 & 88.58 & 62.63 & 61.09 &-\\
        OxfordPet & 31.98 & 60.83 & 61.03 & 58.4 & 87.17 & 83.61 & 87.87 & 94.92 & \cellcolor{gray!30}93.51 & 62.34 & 60.64 &-\\
        Cars & 30.42 & 61.23 & 60.9 & 58.56 & 86.8 & 83.12 & 87.78 & 94.72 & 93.32 & \cellcolor{gray!30}71.25 & 60.22 &-\\
        SUN397 & 31.08 & 59.62 & 62.85 & 57.13 & 86.19 & 81.57 & 87.89 & 94.98 & 93.16 & 70.87 & \cellcolor{gray!30}67.21 & \textbf{72.0} \\

        \hdashline
        \multicolumn{12}{l}{\textbf{Average}} \\
        Average & 33.4 & 60.6 & 60.0 & 54.0 & 75.8 & 76.2 & 87.3 & 59.5 & 90.5 & 65.2 & 61.9 & \textbf{65.9} \\
        \bottomrule
    \end{tabular}
      \caption{Accuracy (\%) of our method (Ours) on the X-TAIL benchmark with order-I. Each row represents the performance on every dataset of the model trained after the corresponding task. \textit{Transfer}, \textit{Average}, and \textit{Last} metrics are shown with bold style.}
    \label{tab:11}
\end{table*}

\end{document}